%% file: main.tex
\documentclass[9.5pt,journal,compsoc]{IEEEtran}

\usepackage[hidelinks,breaklinks=true,bookmarks=false,linkcolor={black},citecolor={black},urlcolor={red}]{hyperref}
\usepackage{graphicx}
\usepackage{enumitem}
\usepackage{amsmath}
\usepackage{amssymb}
\usepackage{multirow}
\usepackage{amsthm}
\usepackage{mathrsfs}
\usepackage{booktabs}
\usepackage{bbding}
\usepackage{bbm}
\usepackage{subfigure}
\usepackage{ragged2e}
\usepackage{tabularx} 
\usepackage{hyperref}
\usepackage{svg}
\usepackage{amsmath}
\usepackage{url}
\usepackage{makecell}
\usepackage{arydshln}
\usepackage{placeins}
\usepackage{longtable}

\renewcommand{\paragraph}[1]{\vspace{2mm}\noindent \textbf{#1}}

\definecolor{gray}{HTML}{efefef}
\definecolor{revise}{rgb}{0.0,0.0,0.0}
\ifCLASSOPTIONcompsoc
  \usepackage[nocompress]{cite}
\else
  \usepackage{cite}
\fi

\begin{document}

% \title{Improving Pseudo-Labeling for Point Cloud Segmentation with Scene-Level Annotation}
\title{Unified Multimodal Understanding and Generation Models: Advances, Challenges, and Opportunities}
% \title{Towards High-Quality Pseudo-Label Generation for Indoor Point Cloud Segmentation with Scene-Level Annotation}
\author{Shanshan Zhao\textsuperscript{*},
Xinjie Zhang\textsuperscript{*}, Jintao Guo\textsuperscript{*}, Jiakui Hu, Lunhao Duan, Minghao Fu, Yong Xien Chng, \\
Guo-Hua Wang,
Qing-Guo Chen\textsuperscript{\dag}, Zhao Xu, Weihua Luo, Kaifu Zhang

\IEEEcompsocitemizethanks{
\IEEEcompsocthanksitem Xinjie Zhang is with Alibaba Group and Hong Kong University of Science and Technology.
\IEEEcompsocthanksitem Jintao Guo and Minghao Fu are with Alibaba Group and Nanjing University.
% \IEEEcompsocthanksitem Dacheng Tao is with Nanyang Technological University.
\IEEEcompsocthanksitem Jiakui Hu is with Alibaba Group and Peking University.
\IEEEcompsocthanksitem Yong Xien Chng is with Alibaba Group and Tsinghua University.
\IEEEcompsocthanksitem Shanshan Zhao, Lunhao Duan, Guo-Hua Wang, Qing-Guo Chen, Zhao Xu, Weihua Luo, and Kaifu Zhang are with Alibaba Group.
% \IEEEcompsocthanksitem Dacheng Tao is with Nanyang Technological University.
\IEEEcompsocthanksitem Shanshan Zhao, Xinjie Zhang, and Jintao Guo contributed equally to this work.
\IEEEcompsocthanksitem Project leader: Qing-Guo Chen.
}
%\thanks{(Xinjie Zhang, Jintao Guo, Shanshan Zhao contributed equally to this work; Project leader: Qing-Guo Chen)}
}

\IEEEtitleabstractindextext{%
\justify
% \begin{abstract}
\input{sec/0_abstract}
% \end{abstract}

}

\maketitle
\IEEEdisplaynontitleabstractindextext

\IEEEpeerreviewmaketitle

\input{sec/1_intro}

\input{sec/2_preliminary}

\input{sec/3_unified_model}

\input{sec/4_dataset}
\input{sec/5_benckmarks}

\input{sec/6_challenges}

\input{sec/7_conclusion}

% \section{Data}
% \input{data}

% \section*{Acknowledgements}
\ifCLASSOPTIONcaptionsoff
  \newpage
\fi

\bibliographystyle{IEEEtran}
\bibliography{main}

% \clearpage
% \section{Appendix}
% \input{6_appendix}

\end{document}

%% file: sec/0_abstract.tex
\begin{abstract}
% We present a comprehensive survey of unified multimodal models that integrate both understanding and generation of vision and language within a single architecture. We categorize existing approaches into three paradigms: pure diffusion, autoregressive, and hybrid frameworks. We analyze their designs for modality encoding, cross-modal attention, and training objectives. We review the large-scale datasets and benchmark suites that have driven recent advances, from web-scale image-text corpora to instruction-guided editing collections. We identify key technical challenges, including efficient tokenization of high-resolution inputs, scalable attention mechanisms, and data quality management. Finally, we outline promising opportunities for modular fine-tuning, hardware acceleration, and extension to additional modalities such as audio and video.
Recent years have seen remarkable progress in both multimodal understanding models and image generation models. Despite their respective successes, these two domains have evolved independently, leading to distinct architectural paradigms: While autoregressive-based architectures have dominated multimodal understanding, diffusion-based models have become the cornerstone of image generation. Recently, there has been growing interest in developing unified frameworks that integrate these tasks. The emergence of GPT-4o's new capabilities exemplifies this trend, highlighting the potential for unification. However, the architectural differences between the two domains pose significant challenges.
To provide a clear overview of current efforts toward unification, we present a comprehensive survey aimed at guiding future research. First, we introduce the foundational concepts and recent advancements in multimodal understanding and text-to-image generation models. Next, we review existing unified models, categorizing them into three main architectural paradigms: diffusion-based, autoregressive-based, and hybrid approaches that fuse autoregressive and diffusion mechanisms. For each category, we analyze the structural designs and innovations introduced by related works. Additionally, we compile datasets and benchmarks tailored for unified models, offering resources for future exploration. Finally, we discuss the key challenges facing this nascent field, including tokenization strategy, cross-modal attention, and data.
As this area is still in its early stages, we anticipate rapid advancements and will regularly update this survey. Our goal is to inspire further research and provide a valuable reference for the community. The references associated with this survey are available on \href{https://github.com/AIDC-AI/Awesome-Unified-Multimodal-Models}{https://github.com/AIDC-AI/Awesome-Unified-Multimodal-Models}
\end{abstract}

%% file: sec/1_intro.tex
\section{Introduction}
\label{sec:intro}
\begin{figure*}[tb!]
    \begin{center}
    \includegraphics[width=\linewidth]{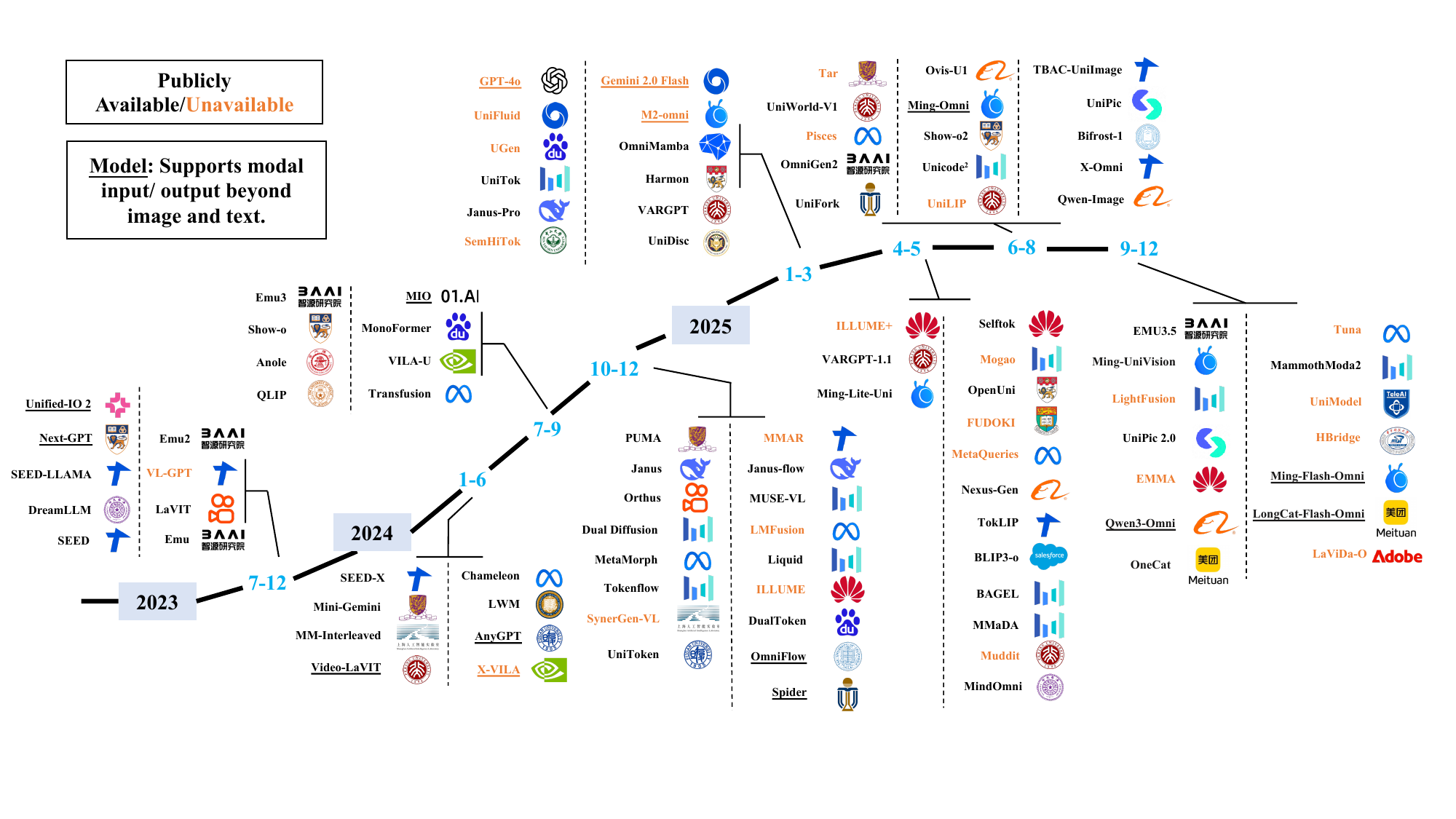}
    \end{center}
    \caption{Timeline of Publicly Available and Unavailable Unified Multimodal Models. The models are categorized by their release years, from 2023 to 2025. Models underlined in the diagram represent \textit{any-to-any multimodal models}, capable of handling inputs or outputs beyond text and image, such as audio, video, and speech. The timeline highlights the rapid growth in this field.}
    \label{fig:overall}
\end{figure*}

In recent years, the rapid advancement of large language models (LLMs), such as LLaMa~\cite{touvron2023llama,touvron2023llama2}, PanGu~\cite{zeng2021pangu,ren2023pangu}, Qwen~\cite{bai2023qwen,yang2024qwen2}, and GPT~\cite{brown2020language}, has revolutionized artificial intelligence. These models have scaled up in both size and capability, enabling breakthroughs across diverse applications. Alongside this progress, LLMs have been extended into multimodal domains, giving rise to powerful multimodal understanding models like LLaVa~\cite{liu2023visual}, Qwen-VL~\cite{wang2024qwen2,bai2025qwen2}, InternVL~\cite{chen2024internvl}, Ovis~\cite{lu2024ovis}, and GPT4~\cite{achiam2023gpt}. These models have expanded their capabilities beyond simple image captioning to performing complex reasoning tasks based on user instructions. On the other hand, image generation technology has also experienced rapid development, with models like SD series~\cite{rombach2022high,esser2024scaling} and FLUX~\cite{flux2024} now capable of producing high-quality images that adhere closely to user prompts.

The predominant architectural paradigm for LLMs and multimodal understanding models is autoregressive generation~\cite{radford2018improving}, which relies on decoder-only structures and next-token prediction for sequential text generation. In contrast, the field of text-to-image generation has evolved along a different trajectory. Initially dominated by Generative Adversarial Networks (GANs)~\cite{goodfellow2020generative}, image generation has since transitioned to diffusion-based models~\cite{ho2020denoising}, which leverage architectures like UNet~\cite{rombach2022high} and DiT~\cite{peebles2023scalable,chen2023pixart} alongside advanced text encoders such as CLIP~\cite{radford2021learning} and T5~\cite{raffel2020exploring}. Despite some explorations into using LLM-inspired architectures for image generation~\cite{sun2024autoregressive,li2024autoregressive,li2024scalable}, diffusion-based approaches remain the state-of-the-art in terms of performance currently.

While autoregressive models lag behind diffusion-based methods in image generation quality, their structural consistency with LLMs makes them particularly appealing for developing unified multimodal systems. A unified model capable of both understanding and generating multimodal content holds immense potential: it could generate images based on complex instructions, reason about visual data, and visualize multimodal analyses through generated outputs. The unveiling of GPT-4o's enhanced capabilities~\cite{openai2024gpt4oblog} in March 2025 has further highlighted this potential, sparking widespread interest in unification.

However, designing such a unified framework presents significant challenges. It requires integrating the strengths of autoregressive models for reasoning and text generation with the robustness of diffusion-based models for high-quality image synthesis. Key questions remain unresolved, including how to tokenize images effectively for autoregressive generation. Some approaches~\cite{fan2025unified,liu2024world,team2024chameleon} employ VAE~\cite{kingma2013auto} or VQ-GAN~\cite{esser2021taming}  commonly used in diffusion-based pipelines, or relevant variants, while others~\cite{sun2024generative,dong2023dreamllm,zhu2023vl} utilize semantic encoders like EVA-CLIP~\cite{sun2023eva} and OpenAI-CLIP~\cite{radford2021learning}. Additionally, while discrete tokens are standard for text in autoregressive models, continuous representations may be more suitable for image tokens, as suggested by emerging research~\cite{li2024autoregressive}. Beyond tokenization, hybrid architectures~\cite{zhao2024monoformer,zhou2024transfusion,xie2024show} that combine parallel diffusion strategies with sequential autoregressive generation offer another promising approach aside from naive autoregressive architecture. Thus, both image tokenization techniques and architectural designs remain in their nascent stages for unified multimodal models.

To provide a comprehensive overview of the current state of unified multimodal models (as illustrated in Fig.~\ref{fig:overall}), thereby benefiting future research endeavors, we present this survey. We begin by introducing the foundational concepts and recent advancements in both multimodal understanding and image generation, covering both autoregressive and diffusion-based paradigms. Next, we review existing unified models, categorizing them into three main architectural paradigms: diffusion-based, autoregressive-based, and hybrid approaches that fuse autoregressive and diffusion mechanisms. Within the autoregressive and hybrid categories, we further classify models based on their image tokenization strategies, reflecting the diversity of approaches in this area.

Beyond architecture, we assemble datasets and benchmarks tailored for training and evaluating unified multimodal models. These resources span multimodal understanding, text-to-image generation, image editing, and other relevant tasks, providing a foundation for future exploration. Finally, we discuss the key challenges facing this nascent field, including efficient tokenization strategy, data construction, model evaluation, etc. Tackling these challenges will be crucial for advancing the capabilities and scalability of unified multimodal models.

In the community, there exist excellent surveys on large language models \cite{zhao2023survey,chang2024survey}, multimodal understanding \cite{liang2024survey,cui2024survey,yin2024survey}, and image generation~\cite{croitoru2023diffusion,bie2024renaissance}, while our work focuses specifically on the integration of understanding and generation tasks. Readers are encouraged to consult these complementary surveys for a broader perspective on related topics.
We aim to inspire further research in this rapidly evolving field and provide a valuable reference for the community. Materials including relevant references, datasets, and benchmarks associated with this survey are available on \href{https://github.com/AIDC-AI/Awesome-Unified-Multimodal-Models}{GitHub} and will be regularly updated to reflect ongoing advancements.

%% file: sec/2_preliminary.tex
\section{Preliminary}
\label{sec:preliminary}
\begin{figure}[tb!]
    \begin{center}
    \includegraphics[width=\linewidth]{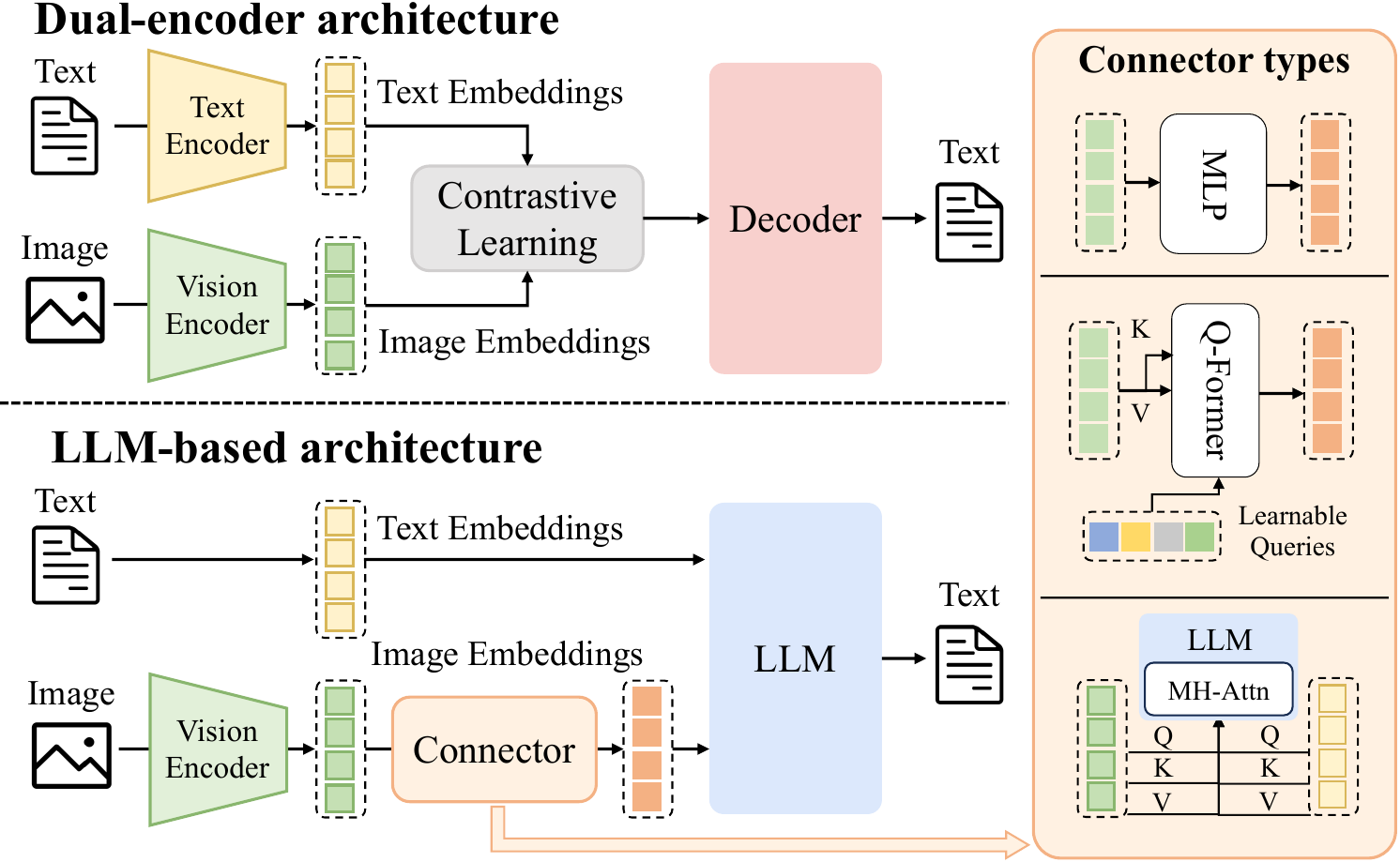}
    \end{center}
    \caption{Architecture of multimodal understanding models, containing multimodal encoders, a connector, and a LLM. The multimodal encoders transform images, audio, or videos into features, which are processed by the connector as the input of LLM. The architectures of theconnector can be broadly categorized by three types: projection-based, query-based, and fusion-based connectors.}
    \label{Fig. understanding}
\end{figure}

\begin{figure}[tb!]
    \begin{center}
    \includegraphics[width=\linewidth]{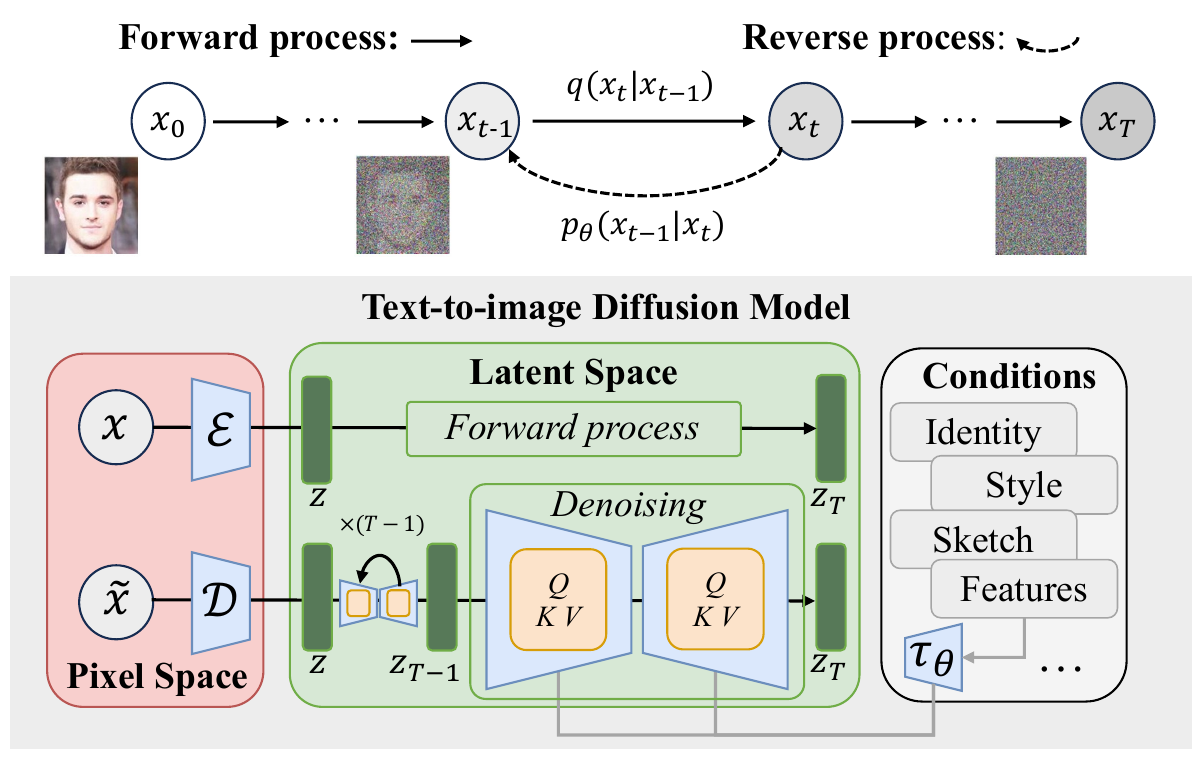}
    \end{center}
    \caption{Illustration of diffusion-based text-to-image generation models, where various conditions beyond text are introduced to steer the outcomes. The image generation is formulated as a pair of Markov chains: a forward process that gradually corrupts input data by adding Gaussian noise, and a reverse process that learns a parameterized distribution to iteratively denoise back to the input data.
    }
    \label{Fig. diffusion}
\end{figure}

\begin{figure}[tb!]
    \begin{center}
    \includegraphics[width=\linewidth]{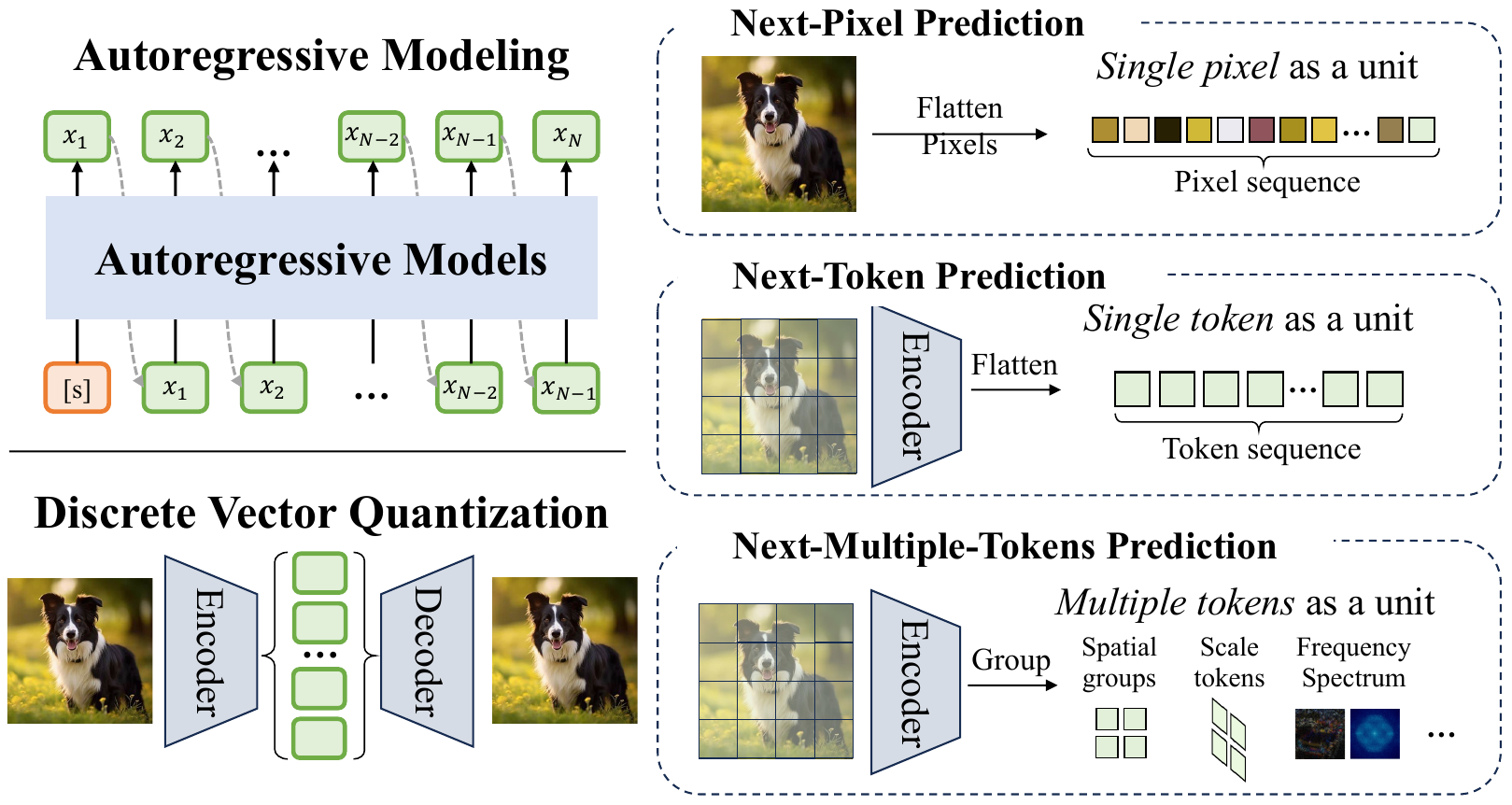}
    \end{center}
    \caption{Illustration of core components in autoregressive models, including the autoregression sequence modeling and discrete vector quantization. Exiting autoregressive models can be roughly divided into three types: Next-Pixel Prediction flattens the image into a pixel sequence, Next-Token Prediction converts the image into a token sequence via a visual tokenizer, and Next-Multiple-Tokens Prediction outputs multiple tokens in an autoregressive step.
    }
    \label{Fig. AR}
\end{figure}

\subsection{Multimodal Understanding Model}
Multimodal understanding models refer to LLM-based architectures capable of receiving, reasoning over, and generating outputs from multimodal inputs \cite{10.1093/nsr/nwae403}. These models extend the generative and reasoning capabilities of LLMs beyond textual data, enabling rich semantic understanding across diverse information modalities \cite{carolan2024review, liang2024survey}.
Most efforts of existing methods focus on vision-language understanding (VLU), which integrates both visual (\textit{e.g.}, images and videos) and textual inputs to support a more comprehensive understanding of spatial relationships, objects, scenes, and abstract concepts \cite{bordes2024introduction, hartsock2024vision, lisurvey}.
A typical architecture of multimodal understanding models is illustrated in Fig.~\ref{Fig. understanding}. 
These models operate within a hybrid input space, where textual data are represented discretely, while visual signals are encoded as continuous representations \cite{li2024continuous}. Similar to traditional LLMs, their outputs are generated as discrete tokens derived from internal representations, using classification-based language modeling and task-specific decoding strategies \cite{li2023blip, liu2023visual}.

Early VLU models primarily focused on aligning visual and textual modalities using dual-encoder architectures, wherein images and text are first encoded separately and then jointly reasoned over via aligned latent representations, including CLIP \cite{radford2021learning}, ViLBERT \cite{lu2019vilbert}, VisualBERT \cite{li2019visualbert}, and UNITER \cite{chen2020uniter}. 
Although these pioneering models established key principles for multimodal reasoning, they depended heavily on region-based visual preprocessing and separate encoders, limiting the scalability and generality of the mode.
With the emergence of powerful LLMs, VLU models have progressively shifted toward decoder-only architectures that incorporate frozen or minimally fine-tuned LLM backbones. 
These methods primarily transform image embeddings through a connector with different structures, as illustrated in Fig.~\ref{Fig. understanding}. 
Specifically, MiniGPT-4 \cite{zhu2023minigpt} utilized a single learnable layer to project CLIP-derived image embeddings into the token space of Vicuna \cite{chiang2023vicuna}.
BLIP-2 \cite{li2023blip} introduced a querying transformer, to bridge a frozen visual encoder with a frozen LLM (e.g., Flan-T5 \cite{chung2024scaling} or Vicuna \cite{chiang2023vicuna}), enabling efficient vision-language alignment with significantly fewer trainable parameters. 
Flamingo \cite{alayrac2022flamingo} employed gated cross-attention layers to connect a pretrained vision encoder with a frozen Chinchilla \cite{hoffmann2022training} decoder.
% , achieving strong few-shot performance on visual question answering and image captioning tasks. 

% compared to subsequent unified and decoder-only frameworks.
% Specifically, CLIP \cite{radford2021learning} was trained on 400 million image-text pairs using a contrastive learning objective, enabling effective cross-modal retrieval and zero-shot classification.
% ViLBERT \cite{lu2019vilbert} and VisualBERT \cite{li2019visualbert} introduced cross-attention mechanisms between visual region features and textual tokens, substantially improving performance on visual understanding tasks. 
% UNITER \cite{chen2020uniter} further advanced this paradigm by integrating object-level visual features, which are extracted through a pre-trained object detector, together with textual tokens within a single Transformer encoder, thereby achieving fine-grained cross-modal alignment at the object level.
% Although these pioneering models established key principles for multimodal reasoning, they depended heavily on region-based visual preprocessing and separate encoders, which limited scalability and generality compared to subsequent unified and decoder-only frameworks.
% , thereby facilitating visually grounded instruction following by minimal additional training. 
% These models highlight a clear design trend: freezing the majority of LLM components while leveraging lightweight modules to integrate vision-language information efficiently.
% like projectors, adapters, or query transformers.

Recent advances in VLU highlight a shift toward general multimodal understanding. 
GPT-4V \cite{openai2023gpt4v} extends the GPT-4 framework \cite{achiam2023gpt} to analyze image inputs provided by the user, demonstrating strong capabilities in visual reasoning, captioning, and multimodal dialogue, despite its proprietary nature. 
Gemini \cite{team2023gemini}, built upon a decoder-only architecture, supports image, video, and audio modalities, with its Ultra variant setting new benchmarks in multimodal reasoning tasks. 
The Qwen series exemplifies scalable multimodal design: Qwen-VL \cite{bai2023qwen} incorporates visual receptors and grounding modules, while Qwen2-VL \cite{wang2024qwen2} adds dynamic resolution handling and M-RoPE for robust processing of varied inputs. 
LLaVA-1.5 \cite{liu2024improved} and LLaVA-Next \cite{liu2024llavanext} use CLIP-based vision encoders and Vicuna-style LLMs for competitive performance in VQA and instruction-following tasks.
The InternVL series \cite{chen2024internvl,chen2024far,chen2024expanding} explore a unified multimodal pre-training strategy,  which simultaneously learns from both text and visual data to enhance performance across various visual-linguistic tasks.
Ovis \cite{lu2024ovis} introduces a structural embedding alignment mechanism through a learnable visual embedding lookup table, thus producing visual embeddings that structurally mirror textual tokens. 
Recently, some models have explored scalable and unified architectures for multimodal processing. 
DeepSeek-VL2 \cite{wu2024deepseek} employs a Mixture-of-Experts (MoE) architecture to enhance cross-modal reasoning.
% TransFusion \cite{zhou2024transfusion} jointly encodes image features and textual tokens within a single Transformer backbone
Overall, these models mark a clear progression toward instruction-tuned and token-centric frameworks capable of addressing diverse multimodal tasks in a unified and scalable manner.

% The Qwen series exemplifies scalable and flexible multimodal design: Qwen-VL \cite{bai2023qwen} incorporates visual receptors, grounding modules, and text-reading capabilities to support a wide range of vision-language tasks, while its successor, Qwen2-VL \cite{wang2024qwen2}, introduces dynamic resolution handling and Multimodal Rotary Position Embedding (M-RoPE) for robust processing of arbitrary image resolutions and extended video inputs.
% LLaVA-1.5 \cite{liu2024improved} and LLaVA-Next \cite{liu2024llavanext} adopt CLIP-based vision encoders coupled with Vicuna-style LLMs through lightweight projection layers, enabling competitive performance in VQA and instruction-following tasks with relatively modest model sizes. 
% Emu2 \cite{sun2024generative} and Emu3 \cite{wang2024emu3} further advance modality unification by encoding images as discrete tokens via a vector-quantized GAN (MoVQGAN \cite{zheng2022movq}), allowing seamless integration of image and text inputs under a shared LLM-based decoder.
% Ovis \cite{lu2024ovis} introduces a structural embedding alignment mechanism through a learnable visual embedding lookup table, thus producing visual embeddings that structurally mirror textual tokens.
% and achieving superior performance over similarly sized multimodal LLMs across various benchmarks.

\subsection{Text-to-Image Model}

% \textbf{Model architectures}
\textit{Diffusion models.} 
Diffusion models (DM) formulate generation as a pair of Markov chains: a forward process that gradually corrupts data $x_{0}$ by adding Gaussian noise over $T$ timesteps to produce $x_T$, and a reverse process that learns a parameterized distribution to iteratively denoise back to the data manifold \cite{sohl2015deep,ho2020denoising,cao2024controllable}. 
Formally, as shown in Fig.~\ref{Fig. diffusion} in the forward process, given the date distribution $x_0 \sim q(x_0)$, at each step $t$, the data $x_t$ is noised:
\begin{equation}
 q(x_{1:T}|x_{0}) := \prod_{t=1}^{T} q(x_t|x_{t-1}), 
\end{equation}
\begin{equation}
 q(x_t|x_{t-1}) = \mathcal{N}(x_t; \sqrt{1 - \beta_t} x_{t-1}, \beta_{t} \textbf{I}), 
\end{equation}
where $\beta_t$ is the variance hyperparameters of the noise.
During the reverse process, the model progressively denoises the data to approximate the reverse of the Markov chain. The reverse transition $p_{\theta}(x_{t-1}|x_{t})$ is parameterized as:
\begin{equation}
 p_{\theta}(x_{t-1}|x_{t}) = \mathcal{N}(x_{t-1}; \mu_{\theta}(x_t, t), \Sigma_{\theta}(x_t, t)), 
\end{equation}
where the network parameterizes the mean $\mu_{\theta}(x_t, t)$ and variance $\Sigma_{\theta}(x_t, t)$. 
The network takes the noised data $x_t$ and time step $t$ as inputs, and outputs the parameters of the normal distribution for noise prediction. 
The noise vector is initiated by sampling $x_T \sim p(x_T)$, and then successively sample from the learned transition kernels $x_{t-1} \sim p_{\theta}(x_{t-1}|x_t)$ until $t=1$. 
The training objective is to minimize a Variational Lower-Bound of the Negative Log-Likelihood: $ \mathcal{L} = \mathbb{E}_{q(x_0, x_{1:T})} \left[ \| \epsilon_{\theta}(x_t, t) - \epsilon^*(x_t, t) \|^2 \right]$, where  $\epsilon_{\theta}(x_t, t)$ is the model's prediction of the noise at timestep $t$, and $\epsilon^*(x_t, t)$ is the true noise added at that timestep.

Early diffusion models utilized a U-Net architecture to approximate the score function \cite{ho2020denoising}. The U-Net design, based on a Wide ResNet, integrates residual connections and self-attention blocks to preserve gradient flow and recover fine-grained image details. 
These methods could be roughly divided into pixel-level methods and latent-feature-level methods. 
The pixel-level methods directly operate the diffusion process in the pixel space, including GLIDE \cite{nichol2022glide} that introduced ``classifier-free guidance'' and Imagen \cite{saharia2022photorealistic} that employ the pretrained large language model, \textit{i.e.}, T5-XXL \cite{raffel2020exploring}, as text encoder.
However, these methods suffer expensive raining and inference computation costs, leading to the development of Latent Diffusion Models (LDMs) \cite{rombach2022high} that operate in the latent space of a pre-trained variational autoencoder. 
LDMs achieve computational efficiency while preserving high-generation quality, thus inspiring various diffusion-based generative models, including VQ-Diffusion \cite{gu2022vector}, SD 2.0 \cite{SD2.0}, SD XL \cite{SD2.1}, and UPainting \cite{li2022upainting}. 
% DALL$\cdot$E 2 \cite{ramesh2022hierarchical} uses a CLIP text encoder coupled with a diffusion prior to map text to image embeddings, ensuring stylistic richness and semantic consistency, 

Advancements in transformer architectures have led to the adoption of transformer-based models in diffusion processes.
The pioneering Diffusion Transformers (DiT) \cite{peebles2023scalable}  transforms input images into a sequence of patches and feeds them through a series of transformer blocks. DiT takes additional conditional information such as the diffusion timestep $t$ and a conditioning signal $c$ as inputs. 
The success of DiT inspired many advanced generative methods, including REPA \cite{yu2024representation} that injects self-supervised visual representations into diffusion training to strengthen large-scale performance, SD 3.0 \cite{esser2024scaling} use two separate sets of weights to model text and image modality, and others \cite{lee2024dit,li2024scalability,chen2024pixart}. 
For text encoders, these methods primarily use utilized contrastive learning to align image and text modalities in a shared latent space, which jointly trained separate image and text encoders on large-scale image-caption pairs \cite{radford2021learning,li2023blip,jia2021scaling}. Specifically, GLIDE \cite{nichol2022glide} explores both CLIP guidance and classifier-free guidance, demonstrating that CLIP-conditioned diffusion outperforms earlier GAN baselines and supports powerful text-driven editing. SD \cite{rombach2022high} employs a frozen CLIP-ViT-L/14 encoder to condition its latent diffusion denoiser, achieving high-quality samples with efficient computation. SD 3.0 \cite{esser2024scaling} utilizes CLIP ViT-L/14, OpenCLIP bigG/14, and T5-v1.1 XXL to transform text into embeddings for generation guidance.

Recent advancements in diffusion models have incorporated LLMs to enhance text-to-image diffusion generation \cite{zhang2024realcompo,wang2025recurrent}, which significantly improves the text-image alignment as well as the quality of generated images.
% RealCompo \cite{zhang2024realcompo} utilizes LLMs to enhance the compositional generation of diffusion models by generating images grounded on bounding box layouts from the LLM. 
RPG \cite{wang2025recurrent} leverages the vision-language prior of multimodal LLMs to reason out complementary spatial layouts from text prompts, and manipulates the object compositions for diffusion models in both text-guided image generation and editing process.
However, these methods require different model architectures, training strategies, and parameter configurations for specific tasks, which presents challenges in managing these models. 
A more scalable solution is to adopt a \textit{unified generation model} capable of handling a variety of data generation tasks \cite{xiao2024omnigen,chen2024unireal,wang2024genartist,fu2025univg}.
OmniGen \cite{xiao2024omnigen} achieves text-to-image generation capabilities and supports various downstream tasks, such as image editing, subject-driven generation, and visual-conditional generation.
UniReal \cite{chen2024unireal} treats image-level tasks as discontinuous video generation, treating varying numbers of input and output images as frames, enabling seamless support for tasks such as image generation, editing, customization, and composition.
GenArtist \cite{wang2024genartist} provides a unified image generation and editing system, coordinated by a multimodal large language model (MLLM) agent.
UniVG \cite{fu2025univg} treats multi-modal inputs as unified conditions with a single set of weights to enable various downstream applications. 
As research in this domain advances, it is expected that increasingly unified models will emerge, capable of addressing a broader spectrum of image generation and editing tasks.

\textit{Autoregressive models.}
Autoregressive (AR) models define the joint distribution of a sequence by factorizing it into a product of conditional probabilities, whereby each element is predicted in turn based on all previously generated elements. This paradigm, originally devised for language modeling, has been successfully adapted to vision by mapping an image to a 1D sequence of discrete tokens (pixels, patches, or latent codes).
Formally, given a sequence $x=(x_1, x_2, ..., x_N)$, the model is trained to generate each element by conditioning all preceding elements:
\begin{equation}
 p(x) = \prod_{i=1}^{N} p(x_i | x_1, x_2, ..., x_{i-1}; \theta).
\end{equation}
where $\theta$ is the model parameters.
The training objective is to minimize the negative log-likelihood(NLL) loss:
\begin{equation}
 \mathcal{L}(\theta) = -\sum_{i=1}^N \log p(x_i | x_1, x_2, ..., x_{i-1}; \theta).
\end{equation}
As shown in Fig.~\ref{Fig. AR}, existing methods are divided into three types based on sequence representation strategies: pixel-based, token-based, and multiple-tokens-based models.

1) Pixel-based models. PixelRNN \cite{van2016pixel} was the pioneering method for next-pixel prediction. It transforms a 2D image into a 1D sequence of pixels and employs LSTM layers to sequentially generate each pixel based on previously generated values. While effective in modeling spatial dependencies, it suffers from high computational costs.
PixelCNN \cite{van2016conditional} introduces dilated convolutions to more efficiently capture long-range pixel dependencies, while PixelCNN++ \cite{salimans2017pixelcnn} leverages a discretized logistic mixture likelihood and architectural refinements to enhance image quality and efficiency.
Some advanced works \cite{reed2017parallel} have also proposed parallelization methods to reduce computational overhead and enable faster generation, particularly for high-resolution images.
% Despite these advances, pixel-level AR models face two major challenges: 1) Quadratic computational complexity with increasing image resolution, and 2) Redundancy due to treating each pixel as an individual prediction unit, limiting scalability.

2) Token-based models.
Inspired by natural language processing paradigms, token-based AR models convert images into compact sequences of discrete tokens, greatly reducing sequence length and enabling high-resolution synthesis.
This process begins with vector quantization (VQ): an encoder-decoder trained with reconstruction and commitment losses learns a compact codebook of latent indices, after which a decoder-only transformer models the conditional distribution over those tokens \cite{van2017neural}. 
Typical VQ models include VQ-VAE-2 \cite{razavi2019generating}, VQGAN \cite{esser2021taming}, ViT-VQGAN \cite{yu2021vector}, and others \cite{cao2023efficient,yu2024image,zhu2024scaling}
% Typical VQ models including VQ-VAE-2 \cite{razavi2019generating} that constructed hierarchical multi-scale tokenizer, VQGAN \cite{esser2021taming} that replaced pixel-level losses with adversarial objectives, and ViT-VQGAN \cite{yu2021vector} that leverages a vision transformer encoder with normalized embeddings to enlarge the codebook without collapse. Some advanced works have also been proposed to improve the efficiency of the architecture \cite{cao2023efficient,yu2024image} or the utilization of the codebook \cite{zhu2024scaling}.
% Early extensions introduced hierarchical multi-scale tokenizers (VQ-VAE-2) to capture both global layout and fine detail \cite{razavi2019generating} and replaced pixel-level losses with adversarial objectives (VQGAN), yielding state-of-the-art fidelity at million-pixel resolutions \cite{esser2021taming}.
% ViT-VQGAN \cite{yu2021vector} leverages a vision transformer encoder with normalized embeddings to enlarge the codebook without collapse, while Efficient-VQGAN \cite{cao2023efficient} embeds Swin Transformer blocks for localized attention and reduced latency. 
% TiTok \cite{yu2024image} adopts the architectures similar to Q-Former to convert 2D patches into a fixed-length 1D token stream. 
% VQGAN-LC \cite{zhu2024scaling} pre-clusters CLIP patch features into a static codebook of $\sim$100K entries, achieving over 99$\%$ utilization without modifying the original network.
Many works have been investigated to enhance the decoder-only transformer models.
% Based on the above works, researchers have investigated whether the scaling laws observed in LLMs are also applicable to autoregressive vision generation. 
LlamaGen \cite{sun2024autoregressive} applies the VQGAN tokenizer to LLaMA backbones \cite{touvron2023llama,touvron2023llama2}, achieving comparable performance with DiTs and showing that generation quality improves with the increase of parameters.
In parallel, data-efficient variants like DeLVM \cite{guo2024data} achieve comparable fidelity with substantially less data, and models such as AiM \cite{li2024scalable}, ZigMa \cite{hu2024zigma}, and DiM \cite{teng2024dim} integrate linear or gated attention layers from Mamba \cite{gu2023mamba} to deliver faster inference and superior performance.
To enrich contextual modeling, stochastic and hybrid decoding strategies have been proposed. Methods like SAIM \cite{qi2023exploring}, RandAR \cite{pang2024randar}, and RAR \cite{yu2024randomized} randomly permute patch predictions to overcome rigid raster biases, while SAR \cite{liu2024customize} generalizes causal learning to arbitrary orders and skip intervals. 
Hybrid frameworks further blend paradigms: RAL \cite{ak2020incorporating} uses adversarial policy gradients to mitigate exposure bias, ImageBART \cite{esser2021imagebart} interleaves hierarchical diffusion updates with AR decoding, and DisCo-Diff \cite{xu2024disco} augments diffusion decoders with discrete latent for best-in-class FID.

3) Multiple-tokens-based methods. 
To improve generation efficiency, recent AR models have shifted from generating individual tokens to predicting multiple tokens as a group, achieving significant speedups without quality loss. 
Next Patch Prediction (NPP) \cite{pang2024next} aggregates image tokens into patch-level tokens with high information density, thus significantly reducing sequence length.
Similarly, Next Block Prediction (NBP) \cite{ren2025next} extends grouping to large spatial blocks, such as rows or entire frames. 
Neighboring AR (NAR) \cite{he2025neighboring} proposes to predict outward using a localized ``next-neighbor'' mechanism, and Parallel Autoregression (PAR) \cite{wang2024parallelized} partitions tokens into disjoint subsets for concurrent decoding.
MAR \cite{li2024autoregressive} abandons discrete tokenization and fixed ordering in favor of continuous representations trained with a diffusion loss.
Beyond spatial grouping, VAR \cite{tian2024visual} introduced a coarse-to-fine next-scale paradigm, which inspired various advanced methods, including FlowAR \cite{ren2024flowar}, M-VAR \cite{ren2024m}, FastVAR \cite{guo2025fastvar}, and FlexVAR \cite{jiao2025flexvar}.
% : FlowAR \cite{ren2024flowar} replaced the discrete tokenizer with a continuous one and applied flow matching at each resolution; 
% M-VAR \cite{ren2024m} added Mamba-style linear-complexity mechanisms for efficient inter-scale modeling; 
% FastVAR \cite{guo2025fastvar} pruned cached tokens at coarser levels, compensating with earlier feature maps; 
% and FlexVAR \cite{jiao2025flexvar} directly forecasts ground-truth tokens, rather than residual features, to flexibly accommodate arbitrary resolutions and aspect ratios.
% Extending VAR to conditional synthesis, STAR \cite{ma2024star} and VAR-CLIP \cite{zhang2024var} integrate pretrained text encoders and CLIP embeddings via cross-attention to guide multi-scale generation, and ControlVAR \cite{li2024controlvar} jointly models image and pixel-level controls—using teacher-forcing to enable fine-grained controllable outputs.
Some frequency-based methods decompose generation spectrally: FAR \cite{yu2025frequency} and NFIG \cite{huang2025nfig} synthesize low-frequency structures before refining high-frequency details.
xAR \cite{ren2025beyond} abstractly unifies autoregressive units, including patches, cells, scales, or entire images, under a single framework.
These multiple-token methods demonstrate the importance of defining appropriate autoregressive units for balancing fidelity, efficiency, and scalability in modern image generation.

% These models are particularly effective in tasks that require maintaining temporal or spatial dependencies, such as object insertion \cite{yun2024generative}, region replacement \cite{xie2024progressive}, and scene extension \cite{avetisyan2024scenescript}.

Control mechanisms have also been integrated into autoregressive decoders for more precise editing. ControlAR \cite{li2024controlar} introduces spatial constraints such as edge maps and depth cues during decoding, allowing fine-grained control over token-level edits. ControlVAR \cite{li2024controlvar} further advances this concept by implementing scale-aware conditioning on image-level features, enhancing coherence and editability. 
CAR \cite{yao2024car} elaborates on a similar concept, focusing on advanced control mechanisms in autoregressive models to enhance the detail and adaptability of visual outputs. 
For complex scenarios involving multiple objects or temporally coherent sequences, Many-to-Many Diffusion (M2M) \cite{shen2024many} adapts the autoregressive framework for multi-frame generation, ensuring semantic and temporal consistency across images. MSGNet \cite{cardenas2021generating} combines VQ-VAE with autoregressive modeling to preserve spatial-semantic alignment across multiple entities in a scene. In the medical domain, MVG \cite{ren2024medical} extends autoregressive image-to-image generation to tasks such as segmentation, synthesis, and denoising by conditioning on paired prompt-image inputs. 
These text-to-image generation AR methods provide the basics of the model architecture and visual modeling methods, effectively advancing research on unified multimodal models for understanding and generation.

%% file: sec/3_unified_model.tex
\begin{table*}[h!]
\fontsize{6pt}{7.2pt}\selectfont 
\centering
\caption{Overview of Unified Multimodal Understanding and Generation Models. This table categorizes models based on their backbone, encoder-decoder architecture, and the specific diffusion or autoregressive models used. It includes information on model, encoder, decoder and the mask used in image generation. The release dates of these models are also provided, highlighting the evolution of multimodal architectures over time.}
%\vspace{-0.3cm}
\begin{tabular}{c|c|c|cc|c|c|c}
\hline
\multirow{2}*{Model}  & \multirow{2}*{Type}  & \multicolumn{5}{c|}{Architecture} &  \multirow{2}*{Date} \\
  \cline{3-7} & & Backbone & Und. Enc. & Gen. Enc. & Gen. Dec. & Mask &  \\
\hline

 \multicolumn{8}{c}{Diffusion Model} \\
 \hline
Dual Diffusion \cite{li2024dual} & a & D-DiT & \multicolumn{2}{c|}{SD-VAE} & SD-VAE  & Bidirect. & 2024-12 \\
{UniDisc} \cite{swerdlow2025unified} & a & DiT & \multicolumn{2}{c|}{MAGVIT-v2} & MAGVIT-v2  & Bidirect. & 2025-03 \\
MMaDA \cite{yang2025mmada} & a & LLaDA & \multicolumn{2}{c|}{MAGVIT-v2} & MAGVIT-v2  & Bidirect. & 2025-05 \\
{FUDOKI} \cite{wang2025fudoki} & a & DeepSeek-LLM & SigLIP & VQGAN & VQGAN  & Bidirect. & 2025-05 \\
{Muddit} \cite{shi2025muddit} & a & Meissonic (MM-DiT) & \multicolumn{2}{c|}{VQGAN} & VQGAN  & Bidirect. & 2025-05 \\
Lavida-O~\cite{li2025lavidao} & a & LaViDa & SigLIP & VQ-Encoder & VQ-Encoder  & Bidirect. & 2025-09 \\
UniModel~\cite{zhang2025unimodel}  & a & MMDiT in Qwen-Image & \multicolumn{2}{c|}{Wan-2.1-VAE} & Wan-2.1-VAE  & Bidirect. & 2025-11 \\
\hline
\multicolumn{8}{c}{Autoregressive Model} \\
\hline
LWM \cite{liu2024world} & b-1 & LLaMa-2 & \multicolumn{2}{c|}{VQGAN} & VQGAN  & Causal  & 2024-02 \\
Chameleon \cite{team2024chameleon} & b-1 & LLaMa-2 & \multicolumn{2}{c|}{VQ-IMG} & VQ-IMG & Causal  & 2024-05 \\
ANOLE \cite{chern2024anole} & b-1 & LLaMa-2 & \multicolumn{2}{c|}{VQ-IMG} & VQ-IMG & Causal & 2024-07 \\
Emu3 \cite{wang2024emu3} & b-1 & LLaMA-2 & \multicolumn{2}{c|}{SBER-MoVQGAN}  & SBER-MoVQGAN & Causal  & 2024-09 \\
MMAR \cite{yang2024mmar}  & b-1 & Qwen2 & \multicolumn{2}{c|}{SD-VAE + EmbeddingViT} & Diffusion MLP & Bidirect. & 2024-10 \\
Orthus \cite{kou2024orthus} & b-1 & Chameleon & \multicolumn{2}{c|}{VQ-IMG+Vision embed.} & Diffusion MLP & Causal  & 2024-11 \\
SynerGen-VL \cite{li2024synergen} & b-1 & InterLM2 & \multicolumn{2}{c|}{SBER-MoVQGAN} & SBER-MoVQGAN & Causal  & 2024-12 \\
Liquid \cite{wu2024liquid}  & b-1& GEMMA & \multicolumn{2}{c|}{VQGAN} & VQGAN & Causal  & 2024-12 \\
UGen \cite{tang2025ugen}  & b-1 & TinyLlama &  \multicolumn{2}{c|}{SBER-MoVQGAN}  & SBER-MoVQGAN & Causal & 2025-03 \\
Harmon \cite{wu2025harmonizing}& b-1 & Qwen2.5 & \multicolumn{2}{c|}{MAR} & MAR & Bidirect.  & 2025-03 \\
TokLIP \cite{lin2025toklip}  & b-1 & Qwen2.5 &  \multicolumn{2}{c|}{VQGAN+SigLIP}  & VQGAN & Causal & 2025-05 \\
Selftok \cite{wang2025selftok}  & b-1 & LLaMA3.1 &  \multicolumn{2}{c|}{SD3-VAE+MMDiT}  & SD3 & Causal & 2025-05 \\
\textcolor{black}{OneCat~\cite{li2025onecat}}& b-1 & Qwen2.5 & Convolution + MLP & VAE in Infinity & VAE in Infinity & Causal & 2025-09 \\
{Emu3.5} \cite{wang2024emu3} & b-1 & Modified Qwen3 & \multicolumn{2}{c|}{SBER-MoVQGAN}  & SBER-MoVQGAN & Causal  & 2025-10 
\\
\hline
Emu \cite{sunemu} & b-2& LLaMA & \multicolumn{2}{c|}{EVA-CLIP} & SD & Causal  & 2023-07 \\
LaVIT \cite{jin2023unified}  & b-2& LLaMA & \multicolumn{2}{c|}{EVA-CLIP} & SD-1.5  & Causal & 2023-09 \\
DreamLLM \cite{dong2023dreamllm} & b-2 & LLaMA & \multicolumn{2}{c|}{OpenAI-CLIP} & SD-2.1 & Causal  & 2023-09 \\
Emu2 \cite{sun2024generative} & b-2& LLaMA & \multicolumn{2}{c|}{EVA-CLIP} & SDXL & Causal  & 2023-12 \\
VL-GPT \cite{zhu2023vl}  & b-2& LLaMA & \multicolumn{2}{c|}{OpenAI-CLIP} & IP-Adapter  & Causal  & 2023-12 \\
MM-Interleaved \cite{tian2024mm}  & b-2& Vicuna  &  \multicolumn{2}{c|}{OpenAI-CLIP} & SD-v2.1  & Causal  & 2024-01 \\
Mini-Gemini \cite{li2024mini} & b-2 & Gemma\&Vicuna  &   \multicolumn{2}{c|}{OpenAI-CLIP+ConvNext} & SDXL  & Causal  & 2024-03 \\
VILA-U \cite{wu2024vila} & b-2& LLaMA-2 & \multicolumn{2}{c|}{SigLIP+RQ} & RQ-VAE & Causal & 2024-09 \\
PUMA \cite{fang2024puma}  & b-2& LLaMA-3  & \multicolumn{2}{c|}{OpenAI-CLIP} & SDXL & Bidirect.& 2024-10 \\
MetaMorph \cite{tong2024metamorph}  & b-2& LLaMA &  \multicolumn{2}{c|}{SigLIP} & SD-1.5 & Causal & 2024-12 \\
ILLUME \cite{wang2024illume}  & b-2& Vicuna & \multicolumn{2}{c|}{UNIT} & SDXL  & Causal  & 2024-12 \\
UniTok \cite{ma2025unitok}  & b-2 & LLaMa-2 & \multicolumn{2}{c|}{ViTamin}  & ViTamin & Causal & 2025-02 \\
{QLIP} \cite{zhao2025qlip} & b-2 &  LlaMa-3 & \multicolumn{2}{c|}{QLIP-ViT+BSQ} & BSQ-AE & Causal & 2025-02\\
% Fair-UMLLM \cite{liu2025fairness} & b-2& LLaMA-2 & \multicolumn{2}{c|}{SigLIP+RQ} & RQVAE & Causal & 2025-02 \\
% ThinkDiff \cite{mi2025think}  & b-2 & Qwen2-VL & \multicolumn{2}{c|}{QwenViT}  & FLUX & Causal & 2025-02 \\
DualToken \cite{song2025dualtoken}  & b-2 & Qwen2.5 & \multicolumn{2}{c|}{SigLIP} & RQVAE  & Causal  & 2025-03 \\
{UniFork} \cite{unifork}  & b-2 & Qwen2.5 & \multicolumn{2}{c|}{SigLIP+RQ} & RQ-VAE & Causal & 2025-06 \\
{UniCode$^2$} \cite{chan2025unicode2}  & b-2& Qwen2.5 & \multicolumn{2}{c|}{SigLIP+RQ} & FLUX.1-dev / SD-1.5  & Causal & 2025-06 \\
{UniWorld} \cite{lin2025uniworld}  & b-2 & Qwen2.5-VL & \multicolumn{2}{c|}{SigLIP2} & DiT & Bidrect. & 2025-06 \\
{Pisces} \cite{xu2025pisces}  & b-2 & LLaMA-3.1 & SigLIP & EVA-CLIP & Diffusion & Causal & 2025-06 \\
{Tar} \cite{han2025tar}  & b-2& Qwen2.5 & \multicolumn{2}{c|}{SigLIP2+VQ} & VQGAN / SANA & Causal & 2025-06 \\
{OmniGen2} \cite{wu2025omnigen2} & b-2 & Qwen2.5-VL & \multicolumn{2}{c|}{SigLIP} & OmniGen & Causal & 2025-06 \\
{Ovis-U1}~\cite{wang2025ovis} & b-2 & Ovis & \multicolumn{2}{c|}{AimV2} & MMDiT & Causal & 2025-06\\
{X-Omni}~\cite{geng2025x} & b-2 &  Qwen2.5-VL & QwenViT & Siglip &  FLUX & Causal & 2025-07\\
{Qwen-Image}~\cite{wu2025qwen} & b-2 & Qwen2.5-VL & \multicolumn{2}{c|}{QwenViT} & MMDiT & Causal & 2025-08 \\
Bifrost-1~\cite{lin2025bifrost}& b-2 & Qwen2.5-VL  & QwenViT & ViT & FLUX & Causal & 2025-08 \\

\textcolor{black}{Ming-UniVision}~\cite{huang2025ming} & b-2 & Ming-UniVision  & \multicolumn{2}{c|}{MingTok}& MingTok & Causal & 2025-10 \\
MammothModa2~\cite{shen2025mammothmoda2}  & b-2 & Qwen3-VL-8B  & QwenViT & MammothTok & Single-stream DiT & Causal & 2025-11 \\
\hline
SEED \cite{ge2023planting} & b-3 & OPT & SEED Tokenizer & Learnable Query & SD  & Causal  & 2023-07 \\
SEED-LLaMA \cite{ge2023making}& b-3 & LLaMa-2 \&Vicuna & SEED Tokenizer & Learnable Query & unCLIP-SD  & Causal& 2023-10 \\
SEED-X \cite{ge2024seed} & b-3&  LLaMa-2 & SEED Tokenizer & Learnable Query & SDXL  & Causal& 2024-04 \\
MetaQueries \cite{pan2025transfer} & b-3 & LLaVA\&Qwen2.5-VL  & SigLIP & Learnable Query & Sana  & Causal  & 2025-04 \\
Nexus-Gen \cite{zhang2025nexus} & b-3 & Qwen2.5-VL  &  QwenViT & Learnable Query & FLUX  & Causal  & 2025-04 \\
Ming-Lite-Uni \cite{Mingunify2025} & b-3 & M2-omni  &  NaViT & Learnable Query & Sana  & Causal  & 2025-05 \\
{BLIP3-o} \cite{chen2025blip3}  & b-3 & Qwen2.5-VL & {OpenAI-CLIP} & Learnable Query  & Lumina-Next & Causal & 2025-05 \\
{OpenUni} \cite{wu2025openuni} & b-3 & InternVL3 & InternViT & Learnable Query & Sana & Causal & 2025-05 \\
% {Ming-Omni} \cite{ai2025mingomni} & b-3 & Ling & QwenViT & Learnable Query & Multi-scale DiT & Causal & 2025-06 \\
{UniLIP}~\cite{tang2025unilip} & b-3 & InternVL3 & InternViT & Learnable Query & Sana & Causal & 2025-07\\
TBAC-UniImage~\cite{xu2025tbac} & b-3 & Qwen2.5-VL & QwenViT & Learnable Query & Sana & Causal & 2025-08 \\
\textcolor{black}{UniPic 2.0~\cite{wei2025skywork}} & b-3 & Qwen2.5-VL & QwenViT & Learnable Query & SD3.5-Medium & Causal & 2025-09 \\
\hline
Janus \cite{wu2024janus}  & b-4& DeepSeek-LLM & SigLIP & VQGAN & VQGAN & Casual &  2024-10 \\ 
Janus-Pro \cite{chen2025janus} & b-4 & DeepSeek-LLM & SigLIP &  VQGAN & VQGAN &  Casual  & 2025-01 \\ 
OmniMamba \cite{zou2025omnimamba} & b-4 & Mamba-2 & DINO-v2+SigLIP &  VQGAN & VQGAN & Causal & 2025-03 \\
Unifluid \cite{fan2024fluid} & b-4 & Gemma-2 & SigLIP & SD-VAE & Diffusion MLP  & Causal  & 2025-03 \\
{MindOmni} \cite{xiao2025mindomni} & b-4 & Qwen2.5-VL & QwenViT & VAE & OmniGen & Causal & 2025-06 \\
{Skywork UniPic}~\cite{wang2025skywork} & b-4 &  Qwen2.5 & SigLIP2 & SDXL-VAE & SDXL-VAE & Causal & 2025-08\\
\hdashline
\rule{0pt}{1.0em}
MUSE-VL \cite{xie2024muse} & b-5 & Qwen-2.5\&Yi-1.5 & SigLIP &  VQGAN & VQGAN & Causal  & 2024-11 \\ 
Tokenflow \cite{qu2024tokenflow}& b-5 & Vicuna\&Qwen-2.5 & OpenAI-CLIP & MSVQ & MSVQ & Causal & 2024-12 \\
VARGPT \cite{zhuang2025vargpt} & b-5 & Vicuna-1.5  & OpenAI-CLIP & MSVQ & VAR-d30  & Causal  & 2025-01 \\
{SemHiTok} \cite{chen2025semhitok}  & b-5 & Qwen2.5 & SigLIP & ViT & ViT & Causal  & 2025-03 \\
VARGPT-1.1 \cite{zhuang2025vargptv1}& b-5 & Qwen2  & SigLIP & MSVQ & Infinity  & Causal  & 2025-04 \\
ILLUME+ \cite{huang2025illume+}  & b-5& Qwen2.5 & QwenViT & MoVQGAN & SDXL  & Causal  & 2025-04 \\
UniToken \cite{jiao2025unitoken}  & b-5& Chameleon & SigLIP & VQ-IMG & VQGAN & Causal  & 2025-04 \\
{Show-o2} \cite{xie2025showo2}  & b-5 & Qwen2.5 & Wan-3DVAE + SigLIP & Wan-3DVAE & Wan-3DVAE & Causal & 2025-06 \\
\hline
 \multicolumn{8}{c}{Fused Autoregressive and Diffusion Model} \\
 \hline
Transfusion \cite{zhou2024transfusion} & c-1 & LLaMA-2 & \multicolumn{2}{c|}{SD-VAE} &SD-VAE  & Bidirect. & 2024-08 \\
Show-o \cite{xie2024show}  & c-1& LLaVA-v1.5-Phi & \multicolumn{2}{c|}{MAGVIT-v2}  & MAGVIT-v2 & Bidirect.  & 2024-08 \\
MonoFormer \cite{zhao2024monoformer} & c-1& TinyLLaMA & \multicolumn{2}{c|}{SD-VAE} & SD-VAE & Bidirect.  & 2024-09 \\
LMFusion \cite{shi2024llamafusion}& c-1 & LLaMA & \multicolumn{2}{c|}{SD-VAE+UNet down.} & SD-VAE+UNet up. & Bidirect. & 2024-12 \\
\textcolor{black}{TUNA~\cite{liu2025tuna}}& c-1 & Qwen-2.5 & \multicolumn{2}{c|}{VAE+Siglip2} & VAE & Bidirect. & 2025-12 \\
\hline
\rule{0pt}{1.0em}
Janus-flow \cite{ma2024janusflow} & c-2 & DeepSeek-LLM & SigLIP & SDXL-VAE & SDXL-VAE & Causal  & 2024-11 \\   
Mogao \cite{liao2025mogao} & c-2 & Qwen2.5 & SigLIP+SDXL-VAE & SDXL-VAE & SDXL-VAE & Bidirect.  & 2025-05 \\  
BAGEL \cite{deng2025bagel} & c-2 & Qwen2.5 & SigLIP & FLUX-VAE & FLUX-VAE & Bidirect.  & 2025-05 \\  
\textcolor{black}{LightFusion~\cite{wang2025lightfusion}}& c-2 & QWen2.5-VL +Wan2.2-TI2V  & QWen2.5-VL & Wan2.2-TI2V & DCAE & Bidirect.  & 2025-11 \\  
HBridge~\cite{wang2025hbridge} & c-2 & QWen + OmniGen2  & QwenViT & SigLip & OmniGen & Bidirect.  & 2025-11 \\  
\textcolor{black}{EMMA~\cite{he2025emma}}& c-2 & Qwen3-34B & SigLIP & DCAE & DCAE & Bidirect.  & 2025-12 \\  
\hline
\end{tabular}
\end{table*}

\begin{figure*}[tb!]
    \begin{center}
    \includegraphics[width=\linewidth]{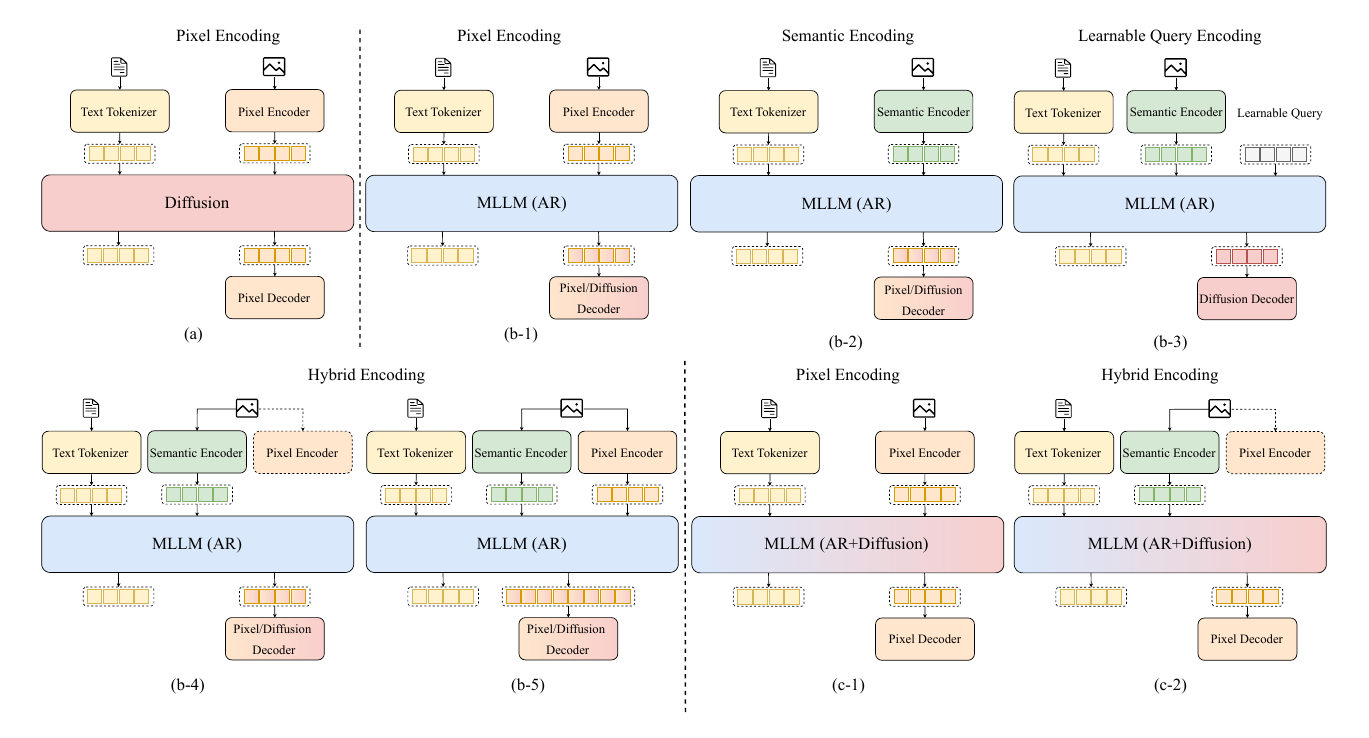}
    \end{center}
    \caption{Classification of Unified Multimodal Understanding and Generation Models. The models are divided into three main categories based on their backbone architecture: Diffusion, MLLM (AR), and MLLM (AR + Diffusion). Each category is further subdivided according to the encoding strategy employed, including Pixel Encoding, Semantic Encoding, Learnable Query Encoding, and Hybrid Encoding. We illustrate the architectural variations within these categories and their corresponding encoder-decoder configurations.}
    \label{fig:unified_model_arch}
\end{figure*}

\begin{table*}[t]
\centering
\scriptsize
\caption{Overview of Any-to-Any Multimodal Models Supporting Modal Input/Output Beyond Image and Text. This table categorizes models that support a variety of input and output modalities, including audio, music, image, video, and text. It includes information on the model's backbone architecture, modality encoders and decoders, the type of attention mask used in vision generation, and the model release dates. These models exemplify the shift toward broader multimodal interactions in recent years.}
\begin{tabular}{c|c|cc|c|c}
\hline
\multirow{2}*{Model}  & \multicolumn{4}{c|}{Architecture} &  \multirow{2}*{Date} \\
  \cline{2-5} & Backbone & Modality Enc. & Modality Dec. & Mask  &  \\
\hline
Next-GPT \cite{wu2024next}  & Vicuna & ImageBind & AudioLDM+SD-1.5+Zeroscope-v2 & Causal & 2023-09 \\
Unified-IO 2 \cite{lu2024unified}  & T5 & Audio Spectrogram Transformer+Vision ViT & Audio ViT-VQGAN + Vision VQGAN & Causal  & 2023-12 \\
Video-LaVIT \cite{jin2024video}  &LLaVA-1.5 & LaVIT+Motion VQ-VAE & SVD img2vid-xt  & Causal & 2024-02 \\
AnyGPT \cite{zhan2024anygpt} & LLaMA-2& Encodec+SEED Tokenizer+SpeechTokenizer & Encodec+SD+SoundStorm & Causal & 2024-02 \\
X-VILA \cite{ye2024x} & Vicuna & ImageBind & AudioLDM+SD-1.5+Zeroscope-v2 & Causal  & 2024-05 \\
MIO \cite{wang2024mio} & Yi-Base & SpeechTokenizer+SEED-Tokenizer & SpeechTokenizer+SEED Tokenizer & Causal  & 2024-09 \\
\multirow{2}*{Spider \cite{lai2024spider}} & \multirow{2}*{LLaMA-2} & \multirow{2}*{ImageBind} & AudioLDM+SD-1.5+Zeroscope-v2 & \multirow{2}*{Causal} & \multirow{2}*{2024-11} \\
& & & +Grounding DINO+SAM& & \\
OmniFlow \cite{li2024omniflow} & MMDiT & HiFiGen+SD-VAE+Flan-T5 & HiFiGen+SD-VAE+TinyLlama & Bidirect. & 2024-12 \\
M2-omni \cite{guo2025m2} & LLaMA-3 & paraformer-zh+NaViT & CosyVoice-vocoder+SD-3 & Casual & 2025-02 \\
% CoDi \cite{tang2023any} & Diff. & Audio+Text+Image+Video Diffusion & 
\hline
\end{tabular}
\end{table*}

\section{Unified Multimodal Models for Understanding and Generation}
Unified multimodal models aim to build a single architecture capable of both understanding and generating data across multiple modalities. These models are designed to process diverse forms of input (e.g., text, image, video, audio) and produce outputs in one or more modalities in a unified manner. A typical unified multimodal framework can be abstracted into three core components: modality-specific encoders that project different input modalities into a representation space; a modality-fusion backbone that integrates information from multiple modalities and enables cross-modal reasoning; and modality-specific decoders that generate output in the desired modality (e.g., text generation or image synthesis). %The overall architecture is illustrated in Fig.~1.

In this section, we primarily focus on unified multimodal models that support vision-language understanding and generation, i.e., models that take both image and text as input and produce either text or image as output. As shown in Fig.~\ref{fig:unified_model_arch}, existing unified models can be broadly categorized into three main types: diffusion models, autoregressive models, and fused AR + diffusion models. For autoregressive models, we further classify them based on their modality encoding methods into four subcategories: pixel-based encoding, semantic-based encoding, learnable query-based encoding, and hybrid encoding. Each of these encoding strategies represents different ways of handling visual and textual data, leading to varying levels of integration and flexibility in the multimodal representations. Fused AR + diffusion models are divided into two subcategories based on modality encoding: pixel-based encoding and hybrid encoding. These models combine aspects of both autoregressive and diffusion techniques, offering a promising approach to more unified and efficient multimodal generation.

In the following sections, we will delve deeper into each category: Section 3.1 explores diffusion-based models, discussing their unique advantages in terms of generating high-quality images and text from noisy representations. Section 3.2 focuses on autoregressive-based models, detailing how different encoding methods impact their performance in vision-language tasks. Section 3.3 covers fused AR + diffusion models, examining how the combination of these two paradigms can enhance multimodal generation capabilities. Finally, we extend our discussion to any-to-any multimodal models, which generalize this framework beyond vision and language to support a broader range of modalities such as audio, video, and speech, with the aim of building universal, general-purpose generative models.

\subsection{Diffusion Models}
Diffusion models have achieved remarkable success in the field of image generation owing to several key advantages. First, they provide superior sample quality compared to generative adversarial networks (GANs), offering better mode coverage and mitigating common issues such as mode collapse and training instability \cite{dhariwal2021diffusion}. Second, the training objective—predicting the added noise from slightly perturbed data—is a simple supervised learning task that avoids adversarial dynamics. Third, diffusion models are highly flexible, allowing the incorporation of various conditioning signals during sampling, such as classifier guidance \cite{dhariwal2021diffusion} and classifier-free guidance \cite{ho2021classifier}, which enhances controllability and generation fidelity. Furthermore, improvements in noise schedules \cite{nichol2021improved} and accelerated sampling techniques \cite{song2021denoising, lu2022dpm} have significantly reduced the computational burden, making diffusion models increasingly efficient and scalable.

Leveraging these strengths, researchers have extended diffusion models beyond unimodal tasks toward multimodal generation, aiming to support both text and image outputs within a unified framework. As shown in Fig.~\ref{fig:unified_model_arch} (a), in multimodal diffusion models, the denoising process is conditioned not only on timestep and noise but also on multimodal contexts, such as textual descriptions, images, or joint embeddings. This extension enables synchronized generation across different modalities and allows for rich semantic alignment between generated outputs.

A representative example is Dual Diffusion \cite{li2024dual}, which introduces a dual-branch diffusion process for joint text and image generation. Specifically, given a text-image pair, Dual Diffusion first encodes the text using a pretrained T5 encoder \cite{raffel2020exploring} with softmax probability modeling to obtain discrete text representations, and encodes the image using the VAE encoder from Stable Diffusion \cite{rombach2022high} to obtain continuous image latents. Both text and image latents are independently noised through separate forward diffusion processes, resulting in noisy latent variables at each timestep.
During the reverse process, the model jointly denoises the text and image latents using two modality-specific denoisers: a Transformer-based text denoiser and a UNet-based image denoiser. Crucially, at each timestep, the denoisers incorporate cross-modal conditioning, where the text latent attends to the image latent and vice versa, enabling semantic alignment between the modalities throughout the denoising trajectory.
After denoising, the text latent is decoded into natural language via a T5 decoder, and the image latent is decoded into a high-fidelity image via the VAE decoder. Training is supervised by two distinct loss terms: the image branch minimizes a standard noise prediction loss, while the text branch minimizes a contrastive log-loss. By coupling the two diffusion chains and introducing explicit cross-modal interactions, Dual Diffusion enables coherent and controllable multimodal generation from pure noise.

{Unlike Dual Diffusion \cite{li2024dual}, which combines discrete text diffusion with continuous image diffusion via Stable Diffusion \cite{rombach2022high}, UniDisc \cite{swerdlow2025unified} employs a fully discrete diffusion framework to train a Diffusion Transformer \cite{Peebles2022DiT} from scratch. It tokenizes text using the LLaMA2 tokenizer \cite{touvron2023llama2} and converts images into discrete tokens with the MAGVIT-v2 encoder \cite{yu2023language}, allowing unification of both modalities in a discrete token space. These tokens undergo a discrete forward diffusion process, where structured noise is added simultaneously across modalities. In the reverse process, UniDisc progressively denoises the tokens to generate coherent sequences. The LLaMA2 and MAGVIT-v2 decoders then transform these sequences into high-quality text and images. By adopting a fully discrete approach, UniDisc enables simultaneous refinement of text and image tokens, enhancing inference efficiency and supporting versatile cross-modal conditioning.}

{In contrast to earlier discrete diffusion-based methods, FUDOKI \cite{wang2025fudoki} introduces a novel generative approach based on a discrete flow matching \cite{gat2024discrete}. Under this framework, FUDOKI models a direct path between noise and data distributions by employing a kinetic-optimal, metric-induced probability trajectory. This design enables a continuous self-correction mechanism, which provides a clear advantage over the simple masking strategies used in earlier models. FUDOKI's model architecture is based on Janus-1.5B \cite{wu2024janus}. However, it introduces essential modifications to support unified vision-language discrete flow modeling. One key change is the replacement of the standard causal mask with a full attention mask. This allows every token to attend to all others, thereby enhancing global contextual understanding. Although this modification removes the explicit causal structure, the model still supports next-token prediction by shifting its output logits by one position. Another important distinction is in the way FUDOKI handles time or corruption levels. Instead of relying on explicit time-step embeddings, as required in diffusion models, FUDOKI infers the corruption state directly from the input data. Following Janus-1.5B, FUDOKI decouples the processing paths for understanding and generation. A SigLIP encoder \cite{zhai2023sigmoid} is employed to capture high-level semantic features for image understanding, while a VQGAN-based tokenizer from LlamaGen \cite{sun2024autoregressive} encodes the image into a sequence of low-level discrete tokens for image generation. At the output stage, the feature embeddings generated by the Janus-1.5B backbone are passed through modality-specific output heads to produce the final text and image outputs.}

{In a similar vein, Muddit \cite{shi2025muddit} introduces a unified model for bidirectional generation using a purely discrete diffusion framework to handle text and images. Its architecture features a single Multimodal Diffusion Transformer (MM-DiT) with an architectural design similar to that of FLUX \cite{blackforestlabs_flux_2024}. To leverage a strong image prior, the MM-DiT generator is initialized from Meissonic \cite{bai2024meissonic}, a model extensively trained for high-resolution synthesis. Both modalities are quantized into a shared discrete space, where a pre-trained VQ-VAE \cite{esser2021taming} encodes images into codebook indices and a CLIP model \cite{radford2021learning} provides text token embeddings. During its unified training, Muddit employs a cosine scheduling strategy to mask tokens, and the single MM-DiT generator is trained to predict the clean tokens conditioned on the other modality. For output, a lightweight linear head decodes text tokens, while the VQ-VAE decoder reconstructs the image, allowing a single set of parameters to handle both text and image generation.}

Building upon this foundation, MMaDA \cite{yang2025mmada} scales up the diffusion paradigm toward a unified multimodal foundation model. It adopts LLaDA-8B-Instruct \cite{nie2025large} as the language backbone and uses a MAGVIT-v2 \cite{yu2023magvit} image tokenizer to convert images into discrete semantic tokens. This unified token space enables seamless multimodal conditioning during generation. To improve alignment across modalities, MMaDA introduces a mixed chain-of-thought (CoT) fine-tuning strategy, which unifies reasoning formats between text and vision tasks. This alignment facilitates cold-start reinforcement learning, allowing effective post-training from the outset. Furthermore, MMaDA incorporates a novel UniGRPO method, a unified policy-gradient-based RL algorithm designed for diffusion models. UniGRPO enables post-training optimization across both reasoning and generation tasks by leveraging diversified reward signals, such as factual correctness, visual-textual alignment, and user preferences. This design ensures the model consistently improves across a broad range of capabilities, rather than overfitting to a narrow task-specific reward. \textcolor{black}{Built on LaViDa~\cite{li2025lavida}, Lavida-O~\cite{li2025lavidao} bridges the gap between unified multi-modal diffusion models and state-of-the-art AR models by implementing a resource-efficient Elastic Mixture-of-Transformers (Elastic-MoT) architecture. To overcome the scarcity of effective masked generative training, it employs progressive upscaling and token compression for robust model scaling. Furthermore, it distinguishes itself through planning and self-reflection mechanisms, which allow the model to use its understanding capability to actively refine and improve generation outputs.}

\textcolor{black}{Building upon the insight from DeepSeek-OCR~\cite{wei2025deepseek} that text can be processed as a visual signal, UniModel~\cite{zhang2025unimodel} unifies these tasks by rendering text into images and employing a diffusion model for both understanding and generation. While this represents a novel strategy, significant challenges remain, including text rendering fidelity, multilingual generation capabilities, and long-context modeling.}

{Despite these innovative approaches, significant challenges and limitations persist in the landscape of unified discrete diffusion models. A primary concern is inference efficiency. Although models like Mercury \cite{labs2025mercuryultrafastlanguagemodels} and Gemini Diffusion \cite{GoogleGeminiDiffusion} demonstrate potential for high-speed parallel token generation, most open-source discrete diffusion models still lag behind the practical inference speeds of their autoregressive counterparts. This discrepancy is primarily due to a lack of support for key-value cache and the degradation in output quality that occurs when decoding multiple tokens in parallel. The effectiveness of diffusion models is also hindered by training difficulties. Unlike autoregressive training, where every token provides a learning signal, discrete diffusion training offers only sparse supervision, as the loss is computed on a randomly selected subset of masked tokens, leading to inefficient use of the training corpus and high variance. Moreover, these models exhibit a length bias and struggle to generalize across different output lengths because they lack a built-in stopping mechanism like the end-of-sequence token found in autoregressive models. Additional development is also needed in architecture and supporting infrastructure. Architecturally, many existing models reuse designs originally created for autoregressive systems, an approach chosen for engineering simplicity that is not always suited to the diffusion process, which aims to capture joint data distributions in a way that is fundamentally different from the sequential nature of autoregressive models. On the infrastructure side, support for discrete diffusion models remains limited. Compared to the mature frameworks available for autoregressive models, they lack well-developed pipelines and robust open-source options. This gap hinders fair comparisons, slows research, and complicates real-world deployment. Addressing these interconnected challenges in inference, training, architecture, and infrastructure is essential to advance the capabilities and practical use of unified discrete diffusion models.}

\subsection{Auto-Regressive Models}
One major direction in unified multimodal understanding and generation models adopts autoregressive (AR) architectures, where both vision and language tokens are typically serialized and modeled sequentially. In these models, a backbone Transformer, typically adapted from large language models (LLMs) such as LLaMA family \cite{touvron2023llama, touvron2023llama2, chen2024allava}, Vicuna \cite{chiang2023vicuna}, Gemma series \cite{team2024gemma, team2024gemma2, team2025gemma}, and Qwen series \cite{bai2023qwen, wang2024qwen2, yang2024qwen2, bai2025qwen2}, serves as the unified modality-fusion module to autoregressively predict multimodal outputs. 

To integrate visual information into the AR framework, as shown in Fig. \ref{fig:unified_model_arch}, existing methods propose different strategies for image tokenization during modality encoding. These approaches can be broadly categorized into four types: pixel-based, semantic-based, learnable query-based, hybrid-based encoding methods.

1) Pixel-based Encoding. As shown in Fig.~\ref{fig:unified_model_arch} (b-1), pixel-based encoding typically refers to the representation of images as continuous or discrete tokens obtained from pretrained autoencoders supervised purely by image reconstruction, such as VQGAN-like models \cite{esser2021taming, zheng2022movq, razzhigaev2023kandinsky, gafni2022make}. These encoders compress the high-dimensional pixel space into a compact latent space, where each spatial patch corresponds to an image token. In unified multimodal autoregressive models, image tokens serialized from such encoders are processed analogously to text tokens, allowing both modalities to be modeled within a single sequence.

Recent works have adopted and enhanced pixel-based tokenization with various encoder designs. 
LWM \cite{liu2024world} employs a VQGAN tokenizer \cite{esser2021taming} to encode images into discrete latent codes without requiring semantic supervision. It proposes a multimodal world modeling framework, wherein visual and textual tokens are serialized together for unified autoregressive modeling. By learning world dynamics purely through reconstruction-based visual tokens and textual descriptions, LWM demonstrates that large-scale multimodal generation is feasible without specialized semantic tokenization.
Both Chameleon \cite{team2024chameleon} and ANOLE \cite{chern2024anole} adopt VQ-IMG \cite{gafni2022make}, an improved VQ-VAE variant designed for content-rich image generation. Compared to standard VQGAN tokenizers, VQ-IMG features a deeper encoder with larger receptive fields and incorporates residual prediction to better preserve complex visual details. This enhancement enables Chameleon and ANOLE to serialize image content more faithfully, thereby supporting high-quality multimodal generation. Moreover, these models facilitate interleaved generation, allowing text and image tokens to be generated alternately within a unified autoregressive framework.
Emu3 \cite{wang2024emu3}, SynerGen-VL \cite{li2024synergen} and UGen \cite{tang2025ugen} employs SBER-MoVQGAN \cite{razzhigaev2023kandinsky, zheng2022movq}, a multi-scale VQGAN variant that encodes images into latent representations capturing both global structure and fine-grained details. By leveraging multi-scale tokenization, these models improve the expressiveness of visual representations for autoregressive modeling while maintaining efficient training throughput. \textcolor{black}{EMU3.5 \cite{cui2025emu3}} develops Discrete Diffusion
Adaptation to accelerate inference via bidirectional parallel prediction. 
Similar with LWM \cite{liu2024world}, Liquid \cite{wu2024liquid} utilizes a VQGAN-style tokenizer and uncovers a novel insight that visual understanding and generation can mutually benefit when unified under a single autoregressive objective and shared visual token representation. Moreover, MMAR \cite{yang2024mmar}, Orthus \cite{kou2024orthus}, Harmon \cite{wu2025harmonizing} introduce the frameworks that utilize continuous-valued image tokens extracted by their corresponding encoders, avoiding the information loss associated with discretization. They also decouple the diffusion process from the AR backbone by employing lightweight diffusion heads atop each auto-regressed image patch embedding. This design ensures that the backbone's hidden representations are not confined to the final denoising step, facilitating better image understanding. TokLIP \cite{lin2025toklip} integrates a low-level discrete VQGAN tokenizer with a ViT-based token encoder SigLIP \cite{zhai2023sigmoid} to capture high-level continuous semantics, which not only empowers visual tokens with high-level semantic understanding but also enhances low-level generative capacity. Selftok \cite{wang2025selftok} introduces a novel discrete visual self-consistency tokenizer,  achieving a favorable trade-off between high-quality reconstruction and compression rate while enabling optimal policy improvement for effective visual reinforcement learning. \textcolor{black}{OneCat~\cite{li2025onecat} employs a patch embedding layer to map raw images into continuous tokens for understanding and editing, alongside a pre-trained multi-scale VAE~\cite{han2024infinity} for reconstruction. Notably, it adopts a hybrid generation strategy: standard next-token prediction for text and next-scale prediction~\cite{han2024infinity} for image synthesis.}

Across most of these models, causal attention masks are applied during both pretraining and generation phases, ensuring that each token only attends to preceding tokens in the sequence. They are trained using a next-token prediction loss, where both image and text tokens are predicted autoregressively, thus unifying the training objective across modalities.
Notably, in pixel-based encoding approaches, the decoder used to reconstruct images from latent tokens typically follows the paired decoder structure originally proposed in VQGAN-like models. These decoders are lightweight convolutional architectures specifically optimized to map discrete latent grids back to the pixel space, focusing primarily on accurate low-level reconstruction rather than high-level semantic reasoning. Moreover, since some methods, like MMAR \cite{yang2024mmar}, Orthus \cite{kou2024orthus} and Harmon \cite{wu2025harmonizing}, tokenize the image into continuous latents, they adopt the lightweight diffusion MLP as their decoder to map continuous latents back to the pixel space.

Despite their effectiveness, pixel-based encoding methods face several inherent limitations:
First, since the visual tokens are optimized purely for pixel-level reconstruction, they often lack high-level semantic abstraction, making cross-modal alignment between text and image representations more challenging. 
Second, pixel-based tokenization tends to produce dense token grids, significantly increasing sequence lengths compared to text-only models, especially for high-resolution images. This leads to substantial computational and memory overhead during autoregressive training and inference, limiting scalability.
Third, because the underlying visual encoders are trained with reconstruction-centric objectives, the resulting visual tokens may retain modality-specific biases, such as excessive sensitivity to textures and low-level patterns, which are not necessarily optimal for semantic understanding or fine-grained cross-modal reasoning.

2) Semantic Encoding. To overcome the semantic limitations inherent in pixel-based encoders, a growing body of work adopts semantic encoding, where image inputs are processed using pretrained text-aligned vision encoders such as OpenAI-CLIP \cite{radford2021learning}, SigLIP \cite{zhai2023sigmoid}, EVA-CLIP \cite{sun2023eva}, or more recent unified tokenizers like UNIT \cite{zhu2024unit}, as shown in Fig.~\ref{fig:unified_model_arch} (b-2). 
{Some of these models leverage the multimodal features encoded by the multimodal autoregressive model as conditions for a diffusion model, enabling image generation with retained multimodal understanding capabilities, like OmniGen2~\cite{wu2025omnigen2} that utilizes Qwen2.5-VL~\cite{bai2025qwen2} as the multimodal model and enhanced OmniGen~\cite{xiao2025omnigen} as the image diffusion model, {Ovis-U1~\cite{wang2025ovis} extending the multimodal model Ovis~\cite{lu2024ovis} into a unified model by incorporating a custom-designed diffusion transformer while Qwen-Image~\cite{wu2025qwen} similarly building upon  Qwen2.5-VL~\cite{bai2025qwen2} by integrating a diffusion transformer}. \textcolor{black}{UniVideo~\cite{wei2025univideo} and Omni-Video~\cite{tan2025omni} extend this paradigm to the video modality.} 
However, most of these models are trained on large-scale image-text pairs with contrastive or regression-based objectives, producing visual embeddings that align closely with language features in a shared semantic space.} Such representations enable more effective cross-modal alignment and are particularly beneficial for multimodal understanding and generation. 

Several representative models leverage different semantic encoders and architectural designs to support unified multimodal tasks.
Emu \cite{sunemu}, Emu2 \cite{sun2024generative}, and LaViT \cite{jin2023unified} all employ EVA-CLIP \cite{sun2023eva} as their vision encoder. Notably, Emu \cite{sunemu} introduces the initial architecture combining a frozen EVA-CLIP encoder, a large language model, and a diffusion decoder to unify VQA, image captioning, and image generation. Emu2 \cite{sun2024generative} builds upon Emu \cite{sunemu} by proposing a simplified and scalable modeling framework for unified multimodal pretraining. It scales the MLLM model up to 37B parameters, significantly enhancing both understanding and generation capabilities. {Bifrost-1~\cite{lin2025bifrost} employs two semantic encoders, ViT for generation and the one employed in the used MLLM (QWen2.5-VL) for understanding. The predicted CLIP latents are used to bridge the MLLM  and diffusion model.} LaViT \cite{jin2023unified} introduces a dynamic visual tokenization mechanism built on top of EVA-CLIP. It employs a selector and merger module to adaptively select visual tokens from image embeddings based on content complexity. This process dynamically determines the length of the visual token sequence per image. The dynamic tokenization significantly reduces redundant information while preserving important visual cues, improving training efficiency and generation quality in tasks such as captioning, visual question answering, and image generation.
DreamLLM \cite{dong2023dreamllm}, VL-GPT \cite{zhu2023vl}, and MM-Interleaved \cite{tian2024mm}, and PUMA \cite{fang2024puma} utilize OpenAI-CLIP encoder \cite{radford2021learning}. DreamLLM \cite{dong2023dreamllm} introduces a lightweight linear projection to align CLIP embeddings with language tokens, while VL-GPT \cite{zhu2023vl} employs a powerful casual transformer after OpenAI-CLIP vision encoder to effectively retain both semantic information and pixel details of the original image. Both MM-Interleaved \cite{tian2024mm} and PUMA \cite{fang2024puma} extract multi-granular image features via a CLIP tokenizer with simple ViT-Adapter or pooling operation to provide fine-grained feature fusion, thus supporting rich multimodal generation. 
Mini-Gemini \cite{li2024mini} introduces a visual token enhancement mechanism that requires dual semantic encoders. Specifically, it leverages a CLIP-pretrained ViT encoder \cite{radford2021learning} to obtain global visual tokens, while a LAION-pretrained ConvNeXt encoder provides dense local visual information. A cross-attention module is then employed to refine the global visual tokens by incorporating detailed visual cues from the dense encoder. These enhanced global tokens are subsequently combined with text tokens and processed by an LLM for joint vision-language understanding and generation. This design effectively bridges the semantic abstraction of CLIP features with the pixel-level precision of dense encoders.
MetaMorph \cite{tong2024metamorph} employs SigLIP \cite{zhai2023sigmoid} to extract visual embeddings and introduces modality-specific adapters within a pretrained language model. These adapters are inserted throughout multiple transformer layers, allowing for deeper vision-language interaction compared to shallow projection approaches.
ILLUME \cite{wang2024illume} adopt UNIT \cite{zhu2024unit} as its vision encoder to provide a unified representation that balances semantic alignment and pixel-level fidelity. Unlike CLIP-like encoders that focus purely on contrastive objectives, UNIT \cite{zhu2024unit} is jointly trained with both image reconstruction and contrastive alignment losses, producing tokens suitable for both vision-language understanding and image synthesis. Built on the powerful UNIT tokenizer, ILLUME effectively generates image tokens that retain both semantic and pixel-level information, which achieves better performance in multiple understanding and generation tasks, including captioning, VQA, text-to-image, and interleaved generation. Similarly, VILA-U \cite{wu2024vila} and UniTok \cite{ma2025unitok} mimic UNIT \cite{zhu2024unit} to introduce image-text contrastive learning to obtain a novel text-aligned vision tokenizer that balances semantic alignment and pixel-level fidelity. {QLIP~\cite{zhao2025qlip} addresses the potential conflict between reconstruction and text-image alignment tasks by implementing binary-spherical quantization. Tar~\cite{han2025tar} initiates the visual codebook by leveraging the vocabulary of LLMs and incorporates scale-adaptive pooling and decoding methodologies. This approach enables the model to adjust the length of the tokenizer according to the requirement: employing coarse-grained tokenizers for efficient generation and fine-grained tokenizers for comprehensive understanding. In generation tasks, Tar utilizes diffusion techniques to enhance the visual generation outcomes of AR models. UniFork~\cite{unifork} capitalizes on the text-aligned vision features of VILA-U. However, differentiating itself from the fully-shared parameters of understanding and generation MLLM, UniFork shares the parameters solely with these tasks at the shallow layer. At the deeper layer, these tasks are managed by distinct networks. This architecture successfully mediates the equilibrium between shared learning and task-specific specialization. UniCode$^2$ \cite{chan2025unicode2} employs a cascaded codebook. In line with the method outlined in \cite{zhuscaling}, it utilizes a substantial codebook derived from clustered SigLIP feature as the frozen foundational codebook, while introducing supplementary learnable codebooks to refine semantics specific to particular tasks. This separation enhances utilization and fosters robust learning.}
Recent work DualToken \cite{song2025dualtoken} {uses} shallow-layer features of SigLIP for reconstruction and deep-layer features of SigLIP for semantic learning, thereby obtaining the texture and semantic visual features simultaneously. As a result, DualToken \cite{song2025dualtoken} achieves superior performance in both reconstruction and semantic tasks while demonstrating remarkable effectiveness in downstream MLLM understanding and generation tasks. {X-Omni~\cite{geng2025x} utilizes SigLIP-VQ as a visual encoder and employs reinforcement learning to mitigate the cumulative error associated with autoregressive inference and to reduce the information loss inherent in discrete encoding. This methodology substantially enhances the generation quality of discrete autoregressive models, facilitating a seamless integration of image and language generation.} \textcolor{black}{Ming-UniVision~\cite{huang2025ming} introduces MingTok, a tokenizer comprising three modules: a low-level encoder, a semantic decoder, and a pixel decoder. The low-level encoder projects the input image into a compact latent space, which is subsequently processed by the semantic decoder for understanding or editing tasks. For generation, predicted tokens within this latent space are reconstructed into raw images via the pixel decoder. MammothModa2~\cite{shen2025mammothmoda2} introduces MammothTok, a unified visual tokenizer built upon AIMv2 for generation, while utilizing QwenViT for understanding. Notably, it leverages multi-layer features from the understanding component to condition the generation process.}  \textcolor{black}{Recently, RAE~\cite{zheng2025diffusion} demonstrated that semantic encoders can also be well adapted to diffusion models, and thus using semantic encoders as visual tokenizers for unified models might become the future direction. VQRAE~\cite{du2025vqrae} has demonstrated that discretized SigLIP features are effective for both understanding and generation tasks. However, the potential for a fully unified framework based on these features remains underexplored..}

Across most of these models, causal attention masks are applied during MLLM training, and next-token prediction loss is used to optimize both text and vision token generation. For image generation, most of these models typically employ diffusion-based decoders, such as SD family \cite{rombach2022high, podell2023sdxl}, IP-adapter \cite{ye2023ip}, FLUX \cite{flux2024}, and Lumina-Next \cite{gao2024lumina-next}, which are trained independently from the MLLM. During inference, the MLLM produces semantic-level visual tokens, which are then passed to the diffusion decoder for final image synthesis. This design choice—pairing semantic encoders with diffusion decoders—is motivated by the fact that semantic embeddings encode high-level conceptual information but lack the spatial density and low-level granularity required for direct pixel reconstruction. Diffusion models, with their iterative denoising mechanisms, are particularly well-suited for this setting: they are capable of progressively refining semantic representations into high-resolution, photorealistic images, even when the input tokens are sparse or abstract. In contrast, although few approaches (i.e., VILA-U \cite{wu2024vila}, UniTok \cite{ma2025unitok}\textcolor{black}{, and VQRAE~\cite{du2025vqrae}}) adopt pixel-based decoders, their generated image quality is less competitive than the diffusion decoders.
Thus, diffusion decoders provide a more robust and expressive decoding pathway for semantically compressed visual tokens, significantly improving text-image alignment, global coherence, and visual fidelity. {UniWorld~\cite{lin2025uniworld} and Pisces~\cite{xu2025pisces} have endeavored to develop and expand such a solution. UniWorld directly utilizes the output features of pre-trained MLLM for visual comprehension as a high-level conditional signal, while employing SigLIP as low-level conditional signal to deliver comprehensive semantic visual control for DiT. Pisces employs EVA-CLIP as a condition for visual generation tasks and leverages diffusion to further enhance the model's visual generation output. For various tasks, Pisces introduces tailored visual vector lengths and employs distinct MLPs to encode conditions. This approach increases the flexibility of model design while mitigating the inference cost compared to a single encoder configuration.} 

Despite these advantages, semantic encoding also comes with several limitations.
First, due to the abstraction of low-level cues, the resulting visual tokens are less controllable at the pixel level, making it difficult to perform fine-grained image editing, local inpainting, or structure-preserving transformation.
Second, semantic encoders often provide only global or mid-level representations, which can be insufficient for tasks requiring spatial correspondence (e.g., referring expression segmentation or pose-accurate synthesis).
Lastly, since the semantic encoder and diffusion decoder are typically trained separately, the lack of end-to-end optimization can lead to mismatch between MLLM outputs and decoder expectations, occasionally causing semantic drift or generation artifacts.

3) Learnable Query Encoding. 
Learnable query encoding has emerged as an effective strategy for producing adaptive and task-relevant image representations. As shown in Fig.~\ref{fig:unified_model_arch} (b-3), instead of relying purely on fixed visual tokenizers or dense image patches, this approach introduces a set of learnable query tokens that dynamically extract informative content from image features. These query tokens act as content-aware probes that interact with visual encoders to generate compact and semantically aligned embeddings, well-suited for multimodal understanding and generation.

Current implementations of learnable query encoding can be broadly divided into two representative paradigms.
The first is represented by SEED \cite{ge2023planting}, which proposes a seed tokenizer that learns causal visual embeddings. Specifically, an input image is first encoded into dense token features via a BLIP-2 ViT encoder \cite{li2023blip}. These features are then concatenated with a set of learnable query tokens and processed by a causal Q-Former to produce causal visual embeddings. This design is trained using both image-text contrastive learning and image reconstruction supervision, allowing the learned embeddings to simultaneously retain low-level visual detail and capture high-level semantic alignment with text. Building on this foundation, SEED-LLAMA \cite{ge2023making} and SEED-X \cite{ge2024seed} enhance the model's capacity by replacing the OPT backbone \cite{zhang2022opt} with a stronger LLaMA2 model \cite{touvron2023llama2} and upgrading the decoder to UnCLIP-SD \cite{rombach2022high} or SDXL \cite{podell2023sdxl}, leading to improved performance in both understanding and generation tasks.
The second approach, introduced by MetaQueries \cite{pan2025transfer}, provides a simplified version of learnable query encoding. Here, image features are extracted via a frozen SigLIP encoder \cite{zhai2023sigmoid}, which are then concatenated with learnable query tokens and directly passed through a frozen vision-language backbone such as LLaVA \cite{chen2024allava} or Qwen2.5-VL \cite{bai2025qwen2}. The output causal embeddings are used as conditioning inputs for a diffusion-based image decoder, enabling high-quality image generation. Because the backbone is kept frozen, the vision-language understanding capabilities remain consistent with the underlying pretrained models, offering a lightweight yet effective solution for multimodal generation. {Open-Uni~\cite{wu2025openuni} refines the architecture of MetaQueries by utilizing solely learnable queries and a lightweight connector between a MLLM and a diffusion model, facilitating cohesive multimodal understanding and generation. Open-Uni demonstrates that the connetor between the MLLM visual understanding component and the diffusion-based visual generation component can be minimal in complexity, exemplified by a configuration comprising merely six Transformer layers.} Nexus-Gen \cite{zhang2025nexus} and Ming-Lite-Uni \cite{Mingunify2025} follow the MetaQueries paradigm, but with notable advancements to further enhance multimodal generation. Nexus-Gen \cite{zhang2025nexus} introduces a more powerful diffusion decoder, FLUX-1.dev, which significantly improves the generation quality. This approach allows the model to better capture the intricate details and high-fidelity features necessary for complex image generation tasks. On the other hand, Ming-Lite-Uni \cite{Mingunify2025} takes a different route by introducing a highly capable MLLM model, M2-omini \cite{guo2025m2}, for enhanced vision-language interaction. This model performs advanced vision-language conditioning to generate the conditioned image embeddings, ensuring a more semantically aligned representation. In addition, Ming-Lite-Uni fine-tunes its diffusion model by incorporating multi-scale learnable tokens, which facilitate improved semantic alignment across various visual scales. The multi-scale representation alignment mechanism enhances the model's ability to generate detailed and contextually rich images from textual prompts, addressing challenges such as resolution mismatches and semantic inconsistencies. This innovative approach makes Ming-Lite-Uni a powerful tool for multimodal understanding and generation, pushing the boundaries of current methods in both flexibility and performance.  {UniLIP~\cite{tang2025unilip} incrementally incorporates reconstruction ability into CLIP through the self-distillation, then employs a learnable query along with the hidden state of the last layer of the MLLM as combined conditions. This framework is demonstrated to optimize the abundant information for visual editing. To exploit the  hierarchical representations within the MLLM’s intermediate layers, TBAC-UniImage~\cite{xu2025tbac} applies learnable queries at multiple layers instead of the last layer.} \textcolor{black}{UniPic 2.0~\cite{wei2025skywork} integrates SD3.5-Medium with Qwen2.5VL-7B via the MetaQuery strategy~\cite{pan2025transfer} and proposes a Progressive Dual-Task
Reinforcement strategy for strengthening both tasks.} 
To sum up, these learnable query-based designs share a common strength: they provide adaptive, compact, and semantically enriched representations that support both efficient image understanding and high-quality generation. By focusing on task-driven token extraction, such models offer a flexible and extensible alternative to traditional visual tokenizers, especially in unified multimodal frameworks.

Despite its flexibility and promising results, learnable query encoding also comes with several limitations that may restrict its broader applicability. 
First, one key challenge is the increased computational overhead introduced by the learnable query tokens. As the number of query tokens grows, the model's memory consumption and computational complexity can significantly rise, especially when scaling up to large datasets or more intricate multimodal tasks. Furthermore, the use of a fixed encoder (as seen in approaches like MetaQueries) can hinder the model's flexibility when confronted with novel or complex visual inputs that diverge from the pretrained data distributions.
Second, in methods like SEED \cite{ge2023planting} and MetaQueries \cite{pan2025transfer}, the reliance on frozen or pretrained backbones can limit the adaptability of visual features to downstream tasks. While freezing reduces training cost and preserves pre-learned knowledge, it also restricts the capacity of the model to dynamically align image features with the evolving query semantics, especially in more diverse or compositional settings.
Finally, while learnable queries effectively capture task-relevant content, they may not always handle diverse visual content uniformly. For instance, complex scenes with multiple objects, fine-grained details, or ambiguous visual cues might not be as well-represented by a relatively small number of learnable queries. This limitation is particularly evident when the model must generate highly detailed outputs, as the fixed or small query set may fail to capture the richness and variability of the visual input in certain contexts.

4) Hybrid Encoding. 
To address the inherent limitations of using a single modality of visual representation, hybrid encoding strategies have been introduced in unified multimodal models. Pixel-based encoding methods (e.g., VQ-VAE or VQGAN) excel at preserving fine-grained visual details but often lack semantic alignment with text. In contrast, semantic-based encoders (e.g., SigLIP or CLIP variants) produce abstract representations that are semantically rich yet less effective at retaining low-level image fidelity. Hybrid encoding aims to combine the strengths of both approaches by incorporating both pixel-level and semantic-level features into a unified representation. Depending on how pixel and semantic tokens are integrated, hybrid encoding methods can be broadly categorized into two types: pseudo hybrid encoding and joint hybrid encoding.

\textit{Pseudo Hybrid Encoding.} 
Representative works in this category include Janus \cite{wu2024janus}, Janus-Pro \cite{chen2025janus}, OmniMamba \cite{zou2025omnimamba}, Unifluid \cite{fan2024fluid}, and MindOmni~\cite{xiao2025mindomni}. As shown in Fig.~\ref{fig:unified_model_arch} (b-4), these models adopt dual encoders—typically a semantic encoder (e.g., SigLIP) and a pixel encoder (e.g., VQGAN or VAE)—but use them in a task-specific manner. During training, the semantic encoder branch is enabled for vision-language understanding tasks, while the pixel encoder branch is activated for image generation tasks. {Although the dual encoders are trained concurrently with combined understanding and generation datasets, the pixel encoder is not utilized during inference in understanding tasks and the semantic encoder is disabled for text-to-image generation. However, for image editing, Unifluid \cite{fan2024fluid} uses the semantic encoder to encode the source image while MindOmni \cite{xiao2025mindomni} utilizes both VAE and semantic encoder to encode the source image.} The rationale behind this design choice is that mixed training with both types of data can enhance performance across understanding and generation tasks. {Skywork UniPic~\cite{wang2025skywork} employs SigLIP2 as the encoder for understanding tasks and MAR~\cite{li2024autoregressive} as the encoder for generative tasks.}  However, since only one encoder is active at any given time, these models do not fully harness the advantages of hybrid encoding. Specifically, they miss the opportunity to employ semantic grounding in generation tasks and fail to utilize high-fidelity visual details in comprehension tasks. Consequently, these models typically engage pixel decoders to reconstruct images from latent codes. 

\textit{Joint Hybrid Encoding.}
As shown in Fig.~\ref{fig:unified_model_arch} (b-5), joint hybrid encoding methods integrate both semantic and pixel tokens into a single unified input for the language model or decoder, enabling simultaneous utilization of both representations. These models differ in their fusion strategies. MUSE-VL \cite{xie2024muse} and UniToken \cite{jiao2025unitoken} concatenates the features from SigLIP and VQGAN along the channel dimension before passing them into the LLM. Tokenflow \cite{qu2024tokenflow} incorporate dual encoders and codebooks with a shared mapping, enabling the joint optimization of high-level semantics and low-level pixel details. VARGPT \cite{zhuang2025vargpt}, VARGPT-1.1 \cite{zhuang2025vargptv1}, and ILLUME+ \cite{huang2025illume+} concatenate the semantic and pixel tokens along the sequence dimension, maintaining both token types in the LLM’s input. {SemHiTok~\cite{chen2025semhitok} introduces the Semantic Guided Hierarchical Codebook (SGHC), which perfectly inherits the semantic information of the semantic codebook while incorporating texture information to achieve pixel reconstruction. It is significant to observe that, contrary to other methods that directly employ distinct network branches for image processing, Show-o2~\cite{xie2025showo2} utilizes separate network branches for the processing of latent features generated by 3DVAE~\cite{wan2025wan}, and uses the spatial-temporal fusion module to aggregate the outputs of different branches. This approach enables Show-o2 to capture both low-level and high-level visual information. However, such an operation might result in the loss of subtle semantic elements, owing to Show-o2's use of 3D VAE for lossy compression of images or videos, potentially causing suboptimal handling of visual semantic details.}
By integrating both semantic and detailed visual information, joint hybrid encoding enables more robust and expressive modeling capabilities for multimodal understanding and generation. These models support pixel decoders (e.g., VQGAN, Infinity \cite{han2024infinity}, VAR-D30 \cite{tian2024visual}) as well as diffusion-based decoders (e.g., SDXL \cite{podell2023sdxl}), allowing them to generate images with improved semantic alignment and visual realism. 

While hybrid encoding offers a promising direction by integrating the complementary strengths of pixel-level and semantic-level representations, it still faces several limitations. Many pseudo hybrid methods do not leverage both encoders simultaneously at inference time, thereby underutilizing the potential synergy between fine-grained visual details and high-level semantics. Even in joint hybrid approaches, the fusion of heterogeneous token types can introduce modality imbalance or redundancy, which may hinder downstream performance if not carefully managed. Additionally, the dual-encoder architecture substantially increases computational and memory overhead, posing challenges for scalability, especially in high-resolution or long-sequence scenarios. Aligning pixel and semantic tokens also remains a non-trivial problem, as implicit mismatches can lead to incoherent representations or conflicting learning signals. Finally, current hybrid encoding techniques often assumes implicit alignment between the pixel and semantic tokens. However, in practice, such alignment is non-trivial. Misalignment between visual details and semantic abstraction can lead to conflicting supervision signals or incoherent representations, especially in data-scarce or noisy training settings.

\subsection{Fused Autoregressive and Diffusion Models}
Fused autoregressive (AR) and diffusion modeling has recently emerged as a powerful framework for unified vision-language generation. In this paradigm, text tokens are generated autoregressively, preserving the compositional reasoning strengths of large language models, while image tokens are generated through a multi-step denoising process, following the diffusion modeling principle. This hybrid strategy allows image generation to proceed in a non-sequential manner, resulting in improved visual quality and global consistency.

Representative models such as Transfusion \cite{zhou2024transfusion}, Show-o \cite{xie2024show}, MonoFormer \cite{zhao2024monoformer}, and LMFusion \cite{shi2024llamafusion}, follow this approach. During generation, noise is added to latent visual representations and removed iteratively, with the process conditioned on previously generated text or full cross-modal context. Although this design increases inference cost due to multiple sampling steps, it achieves an effective trade-off between symbolic control and visual fidelity, making it well-suited for high-quality vision-language generation tasks.
Existing fused AR + diffusion models typically adopt one of two image tokenization strategies: pixel-based encoding and hybrid encoding.

1) Pixel-based Encoding: As shown in Fig.~\ref{fig:unified_model_arch} (c-1), pixel-based encoding transforms images into either discrete tokens or continuous latent vectors, which are then used as targets in a diffusion-based denoising process conditioned on autoregressively generated text tokens. Among recent works, Transfusion \cite{zhou2024transfusion}, MonoFormer \cite{zhao2024monoformer}, and LMFusion \cite{shi2024llamafusion} all adopt continuous latent representations extracted via SD-VAE. These models share a common training objective that combines autoregressive loss for language modeling and diffusion loss for image reconstruction, and utilize bidirectional attention to enable spatial coherence. Despite this shared framework, each model introduces distinct architectural innovations: Transfusion \cite{zhou2024transfusion} proposes a unified transformer backbone with modality-specific layers to jointly handle discrete and continuous inputs; MonoFormer \cite{zhao2024monoformer} introduces a compact architecture with shared blocks and task-dependent attention masking to balance AR and diffusion tasks; and LMFusion \cite{shi2024llamafusion} enables frozen LLMs to perform high-quality image generation through a lightweight visual injection module, preserving language capabilities while training only the vision branch. In contrast, Show-o \cite{xie2024show} employs a discrete pixel-based tokenizer based on MAGVIT-v2 \cite{yu2023magvit}, generating symbolic image tokens compatible with transformer-style decoding. It supports both AR-based text token generation and diffusion-based image synthesis, supervised through a combination of autoregressive and diffusion losses.  \textcolor{black}{TUNA~\cite{liu2025tuna} achieves unified visual representations via a VAE encoder followed with a semantic encoder.} Collectively, these models demonstrate the effectiveness of pixel-based encoding in balancing semantic controllability from language models and high-resolution visual fidelity from diffusion processes. 

Despite their effectiveness, pixel-based encoding approaches in fused AR and diffusion frameworks also face several limitations. First, models that rely on continuous latent spaces (e.g., via SD-VAE) introduce significant computational overhead during training and inference, due to the iterative nature of diffusion sampling and the need for high-dimensional feature processing. This can become especially burdensome when scaling to high-resolution image generation or multi-turn vision-language interactions.
Second, alignment between textual and visual modalities remains challenging. While bidirectional attention mechanisms enable cross-modal fusion, the latent space representations—particularly those learned through unsupervised reconstruction objectives in SD-VAE—may not always be optimally aligned with semantically meaningful language tokens, potentially leading to weaker fine-grained controllability or less interpretable generation.
Finally, discrete tokenization schemes, as used in Show-o, inherit issues from VQ-based models such as codebook collapse and limited capacity to represent subtle visual nuances. These symbolic tokens, while compatible with transformer-style modeling, may constrain visual diversity and reduce reconstruction fidelity compared to continuous latent methods.

2) Hybrid Encoding: As shown in Fig.~\ref{fig:unified_model_arch} (c-2), hybrid encoding fuses both semantic features (e.g., from CLIP or ViT encoders) and pixel-level latents (e.g., from SD-VAE), providing a more expressive image representation. This approach allows models to leverage high-level semantic abstraction while maintaining detailed visual information. 
Specifically, Janus-flow \cite{ma2024janusflow}, Mogao \cite{liao2025mogao} and BAGEL \cite{deng2025bagel} adopt a dual-encoder architecture and presents a minimalist architecture that harmonizes AR language models with rectified flow. They decouples the understanding and generation encoders, using SigLIP or the concatenation of SigLIP and SDXL-VAE as the vision encoder for multimodal understanding and SDXL-VAE or FLUX-VAE for image generation. \textcolor{black}{LightFusion~\cite{wang2025lightfusion} reduces the training difficulty of Bagel by initializing independent understanding and generation models. Omni-View~\cite{hu2025omni} further achieves this unification in multi-view image scenarios by adopting a generative paradigm inspired by MVAR~\cite{hu2025auto}.} \textcolor{black}{EMMA~\cite{he2025emma} concatenates the understanding tokens and generation tokens along the channel dimension  rather than the token-wise concatenation~\cite{deng2025bagel}, enabling the model to handle understanding and generation tasks simultaneously. Meanwhile, EMMA replaces the understanding encoder SigLIP2 with a mixture-of-experts architecture to better handle diverse types of input images. Unlike Bagel, which connects generation and understanding across all layers, HBridge~\cite{wang2025hbridge} is built upon a pre-trained LLM paired with a diffusion-based generative expert, connecting them only via a selective mid-layer bridge.} 
However, the pseudo hybrid encoding design limits the model’s ability to simultaneously leverage both semantic and pixel-level features during generation, as only the pixel encoder is active in the image synthesis process. This decoupling, while beneficial for modularity and training efficiency, prevents the model from fully exploiting semantic cues during image decoding, potentially weakening fine-grained alignment and multimodal compositionality in generative tasks. 

Despite their advancements, hybrid encoding methods face several challenges. The integration of dual-encoder architectures and the combination of autoregressive and diffusion processes increase the model's overall complexity. This can result in higher computational costs and longer training times, making them less efficient compared to simpler models. Furthermore, ensuring effective alignment between semantic and pixel-level features requires careful architectural design and optimization. This alignment process can be difficult to achieve and fine-tune, limiting the model's ability to fully utilize both modalities in a balanced way. Additionally, balancing the objectives of vision-language understanding and image generation within a unified model often leads to trade-offs, where improvements in one task may come at the expense of the other. These limitations underscore the need for more efficient hybrid designs that can better leverage the strengths of both visual and semantic features while reducing computational overhead and maintaining high performance across tasks.

\subsection{Any-to-Any Multimodal Models}
While early unified multimodal models primarily focused on text-image pairs, recent research has expanded toward any-to-any multimodal modeling. This ambitious approach seeks to create models that can process and generate across a diverse set of modalities, including audio, video, speech, music, and beyond. These models aim to unify modality-specific encoders and decoders within a single architecture, enabling tasks such as text-to-audio, video-to-text, speech-to-music, or even image-to-video generation. This section reviews representative works in this emerging field, highlighting their design principles, modularity, and current limitations.

Most any-to-any models follow a modular design, where each modality is paired with a specialized encoder and decoder, while a shared backbone facilitates cross-modal representation learning and sequence modeling. For example, OmniFlow \cite{li2024omniflow} integrates HiFiGen \cite{kong2020hifi} for audio and music generation, SD-VAE \cite{rombach2022high} for image processing, and uses a DiT-like diffusion model (MMDiT) \cite{esser2024scaling} as the backbone. This modular design allows the model to efficiently combine different modalities for complex generation tasks.

Some models rely on shared embedding spaces to unify different modalities at the feature level. For instance, Spider \cite{lai2024spider}, X-VILA \cite{ye2024x}, and Next-GPT \cite{wu2024next} leverage ImageBind—a contrastively trained model that maps six modalities (text, image, video, audio, depth, and thermal) into a single embedding space. This unified representation enables flexible conditioning and generation via modality-specific decoders, such as Stable Diffusion \cite{rombach2022high}, Zeroscope, or LLM-based text decoders \cite{touvron2023llama}. While this approach is elegant in theory, its generative capacity is often constrained by the quality of the decoder and the granularity of the shared embedding.

Other models, such as AnyGPT \cite{zhan2024anygpt} and Unified-IO 2 \cite{lu2024unified}, extend the sequence-to-sequence paradigm to handle multiple modalities. AnyGPT \cite{zhan2024anygpt} utilizes EnCodec \cite{defossez2022high} for audio tokenization, SpeechTokenizer \cite{zhang2023speechtokenizer} for speech, and trains a unified Transformer with modality-specific prefixes. Unified-IO 2 \cite{lu2024unified}, on the other hand, adopts a more structured encoder-decoder design that includes visual, audio, and language modalities, supporting tasks like AST-to-text, speech-to-image, or video captioning within a single model. 

A recent and notable addition to the any-to-any unified multimodal models is M2-omni \cite{guo2025m2}, which introduces a highly versatile architecture capable of processing and generating a wide variety of modalities, including text, image, video, and audio. M2-omini takes a step forward by incorporating multiple modality-specific tokenizers and decoders, each carefully designed to handle the unique characteristics of different data types. Specifically, it utilizes NaViT \cite{dehghani2023patch} to encode videos and images of arbitrary resolution, and combines a pre-trained SD-3 \cite{podell2023sdxl} as the image decoder. For audio, M2-omini introduces paraformer-zh \cite{gao2022paraformer} to extract audio tokens, and feeds the predicted discrete audio tokens into the pretrained CosyVoice \cite{du2024cosyvoice} flow matching and vocoder model to generate audio streams. 
This integration ensures that M2-omini can effectively generate high-quality images, and audio streams from various inputs, making it a truly multi-modal powerhouse. \textcolor{black}{{Ming-Omni~\cite{ai2025mingomni} adheres to the integrated MoE architecture, wherein modality-specific routing is facilitated through dedicated mechanisms tailored for each token, thereby enabling customized routing distributions. To address the multi-scale phenomenon inherent in visual generation~\cite{tian2024visual}, Ming-Omni employs multi-scale learnable queries, directed by an alignment strategy, to iteratively generate images progressing from coarse to fine detail. Furthermore, Ming-Omni integrates the audio modality and implements a dual-stage training strategy to mitigate the mutual influence between audio comprehension and generation tasks. The initial stage emphasizes comprehension capabilities, while the subsequent stage concentrates on enhancing generation quality. BLIP3-o \cite{chen2025blip3} also employs learnable queries to bridge multimodal understanding and generation. However, it utilizes two diffusion models: one for learning CLIP embeddings and the other for using CLIP as a condition to generate images. It reveals that flow matching loss is more effective than MSE loss, enabling more diverse image sampling and yielding better image quality. By using Ling-Flash-2.0~\cite{inclusionAI2025lingflash} as the main backbone, Ming-Falsh-Omni~\cite{ai2025ming} achieves a favorable trade-off between
performance and efficiency. Addressing the specific demands of real-time interaction, Qwen3-Omni~\cite{Qwen3-Omni} proposes a Thinker-Talker MoE architecture, which decouples deep reasoning from streaming speech generation to minimize latency without sacrificing intelligence. Furthermore, LongCat-Flash-Omni~\cite{team2025longcat} extends these capabilities into long-context scenarios (up to 128k tokens), utilizing a Shortcut-connected MoE and curriculum learning strategies to effectively model long-range dependencies in video and audio sequences.}}

Despite promising progress, current any-to-any models still face several challenges. One key issue is modality imbalance, where text and image modalities are often dominant, while others like audio, video, and music are underrepresented. This limits the diversity of tasks these models can handle. Another challenge is scalability, as supporting a wide range of modalities increases model complexity, leading to higher inference latency and greater resource requirements. Additionally, ensuring semantic consistency across modalities remains a non-trivial task, with models often struggling to maintain grounded and aligned outputs. These challenges represent ongoing areas of research in the development of any-to-any multimodal models.

Nevertheless, these models represent a crucial step toward developing universal foundation models that can understand and generate across the full spectrum of human sensory input and communication. As data, architectures, and training paradigms evolve, future any-to-any models are expected to become more compositional, efficient, and capable of truly universal cross-modal generation.

%% file: sec/4_dataset.tex
\section{Datasets on Unified Models}

\begin{table}[t!]
\centering
\scriptsize
\caption{
% Common datasets for pre-training unified multimodal understanding and generation models, including sample counts and release dates, categorized by primary application. 
Overview of common datasets used for pre-training unified multimodal understanding and generation models. This table categorizes datasets by primary application (Multimodal Understanding, Text-to-Image Generation, Image Editing, Interleaved Image-Text, and Other conditional generation tasks), detailing the approximate sample size and release date for each dataset.
}
\begin{tabular}{c|c|c}
\hline
{Dataset} & {Samples} & {Date} \\ 
\hline
\multicolumn{3}{c}{Multimodal Understanding} \\
\hline

    RedCaps~\cite{desai2021redcaps} & 12M & 2021-11 \\
    Wukong~\cite{gu2022wukong} & 100M & 2022-02 \\
    LAION~\cite{schuhmann2022laion} & 5.9B & 2022-03 \\
    COYO~\cite{byeon2022coyo} & 747M & 2022-08 \\
    Laion-COCO~\cite{schuhmann2022laioncoco} & 600M & 2022-09 \\
    DataComp~\cite{gadre2023datacomp} & 1.4B & 2023-04 \\
    %SAM~\cite{kirillov2023segment} & 11M & 2023-04 \\
    GRIT~\cite{peng2023kosmos} & 20M & 2023-06 \\
    CapsFusion-120M~\cite{yu2024capsfusion} & 120M & 2023-10\\
    ShareGPT4V~\cite{chen2024sharegpt4v} & 100K & 2023-11 \\
    {ALLaVA-4V}~\cite{chen2024allava} & 1.4M & 2024-02 \\
    Cambrian-10M(7M)~\cite{tong2024cambrian} & 10M & 2024-06 \\
    LLaVA-OneVision~\cite{li2024llava} & 4.8M & 2024-08 \\
    Infinity-MM~\cite{gu2024infinity} & 40M & 2024-10 \\
    Honey-Data-15M~\cite{zhang2025beehighqualitycorpusfullstack} & 15M & 2025-10 \\
    
\hline
    \multicolumn{3}{c}{Text-to-Image} \\
\hline
    CC-12M~\cite{changpinyo2021conceptual} & 12M & 2021-02 \\
    LAION-Aesthetics~\cite{schuhmann2022laion} & 120M & 2022-08 \\
    SAM~\cite{kirillov2023segment} & 11M & 2023-04 \\
    Mario-10M~\cite{chen2024textdiffuser} & 10M & 2023-05 \\
    RenderedText~\cite{RenderedText} & 12M & 2023-06 \\
    JourneyDB~\cite{sun2023journeydb} & 4M & 2023-07 \\
    AnyWord-3M~\cite{tuo2023anytext} & 3M & 2023-11 \\   
    CosmicMan-HQ 1.0~\cite{li2024cosmicman} & 6M & 2024-04 \\
    {DOCCI}~\cite{onoe2024docci} & 15K & 2024-04 \\
    PixelProse~\cite{singla2024pixels} & 16M & 2024-06 \\
    DenseFusion~\cite{li2024densefusion} & 1M & 2024-07 \\
    Megalith~\cite{BoerBohan2024Megalith10m} & 10M & 2024-07 \\
    text-to-image-2M~\cite{text-to-image-2M} & 2M & 2024-09 \\ 
    PD12M~\cite{meyer2024public} & 12M & 2024-10 \\
    SFHQ-T2I~\cite{david_beniaguev_2024_SFHQ_T2I} & 122K & 2024-10 \\
    EliGen TrainSet~\cite{zhang2025eligen} & 500k & 2025-01 \\
    TextAtlas5M~\cite{wang2025textatlas5m} & 5M & 2025-02 \\
    BLIP-3o 60k~\cite{chen2025blip3} & 60K & 2025-05 \\ 
    {ShareGPT-4o-Image}~\cite{sharegpt4oimage} & 45K & 2025-06 \\ 
    Poster100K~\cite{chen2025postercraft} & 100K & 2025-06 \\
Text-Render-2M~\cite{chen2025postercraft} & 2M & 2025-06 \\
    {Echo-4o-Image}~\cite{ye2025echo} & 106K & 2025-08 \\
    FLUX-Reason-6M~\cite{fang2025flux} & 6M & 2025-09 \\
\hline

    \multicolumn{3}{c}{Image Editing} \\
\hline
    InstructP2P~\cite{brooks2023instructpix2pix} & 313K & 2022-11 \\ 
    Magicbrush~\cite{zhang2023magicbrush} & 10K & 2023-06 \\ 
    {HIVE}~\cite{zhang2023hive} & 1.1M & 2023-07 \\ 
    HQ-Edit~\cite{hui2024hq} & 197K & 2024-04 \\ 
    SEED-Data-Edit~\cite{ge2024seed} & 3.7M & 2024-05 \\ 
    {EditWorld}~\cite{yang2024editworld} & 8.6K & 2024-06 \\
    UltraEdit~\cite{zhao2024ultraedit} & 4M & 2024-07 \\ 
    {PromptFix}~\cite{zeng2024promptfix} & 1M & 2024-09 \\ 
    OmniEdit~\cite{wei2024omniedit} & 1.2M & 2024-11 \\  
    AnyEdit~\cite{yu2024anyedit} & 2.5M & 2024-11 \\ 
    {RefEdit}~\cite{pathiraja2025refedit} & 18K & 2025-04 \\ 
    {Imgedit}~\cite{ye2025imgedit}  & 1.2M & 2025-05 \\ 
    {ByteMorph-6M}~\cite{chang2025bytemorph} & 6.4M & 2025-05 \\ 
    {ShareGPT-4o-Image}~\cite{sharegpt4oimage} & 46K & 2025-06 \\
    {GPT-Image-Edit-1.5M}~\cite{wang2025gpt} & 1.5M & 2025-07 \\
    {X2Edit}~\cite{ma2025x2edit} & 3.7M & 2025-08 \\
    Pico-Banana-400K~\cite{qian2025picobanana400klargescaledatasettextguided} & 400K & 2025-10 \\
\hline
    \multicolumn{3}{c}{Interleaved Image-Text} \\
\hline
    Multimodal C4~\cite{zhu2023multimodal} & 101.2M & 2023-04 \\
    OBELICS~\cite{laurenccon2023obelics} & 141M & 2023-06 \\
    CoMM~\cite{chen2024comm} & 227K & 2024-06 \\
    {OmniCorpus}~\cite{li2024omnicorpus} & 8B & 2024-10 \\
    
\hline
    \multicolumn{3}{c}{Other Text+Image-to-Image} \\
\hline
    LAION-Face~\cite{zheng2022general} & 50M & 2021-12 \\
    MultiGen-20M~\cite{qin2023unicontrol} & 20M & 2023-05 \\ 
    Subjects200K~\cite{tan2024ominicontrol} & 200K & 2024-11 \\ 
    X2I-subject-driven~\cite{xiao2024omnigen} & 2.5M & 2024-12 \\
    SynCD~\cite{kumari2025syncd} & 95K & 2025-02 \\
    {Graph200K}~\cite{li2025visualcloze} & 200K & 2025-03 \\
    
MetaQuery\_Instruct\_2.4M~\cite{pan2025transfer} & 2.4M & 2025-06\\
    {Echo-4o-Image}~\cite{ye2025echo} & 73K & 2025-08 \\
    
\hline
\end{tabular}
\label{tab:datasets}
\end{table}

Large-scale, high-quality, and diverse training data form the bedrock for building powerful unified multimodal understanding and generation models. These models typically require pre-training on vast amounts of image-text pairs to learn cross-modal correlations and representations. It is important to note that before being trained on large-scale multi-modal data, these models are ofter initialized with parameters derived from training on a large-scale natural language corpus, such as Common Crawl~\footnote{https://commoncrawl.org}, RedPajama~\cite{weber2024redpajama}, WebText~\cite{radford2019language}, etc. Since this survey primarily focuses on multimodal models, the discussion in this section will exclude text-only data. Based on the primary use and modality characteristics, common pre-training multimodal datasets can be broadly categorized as follows: Multimodal Understanding datasets, Text-to-Image Generation datasets, Image Editing datasets, Interleaved Image-Text datasets, and other datasets for image generation conditioned on both text and image inputs. This section will elaborate on representative datasets listed in Tab.~\ref{tab:datasets} within each category, focusing on those released from 2020 onwards.

\subsection{Multimodal Understanding Datasets}

These datasets are primarily used to train the cross-modal understanding capabilities of models, enabling tasks such as image captioning, visual question answering (VQA), image-text retrieval, and visual grounding. They typically consist of large collections of images paired with corresponding textual descriptions.

\begin{itemize}
    \item RedCaps~\cite{desai2021redcaps}: This dataset comprises 12 million image-text pairs sourced from Reddit. It is particularly specialized in capturing everyday items and moments (like pets, hobbies, food, leisure, etc.) frequently shared by users on social media platforms.

    \item Wukong~\cite{gu2022wukong}: The Wukong dataset is a large-scale Chinese multimodal pre-training dataset containing 100 million Chinese image-text pairs filtered from the web. Its creation addressed the lack of large-scale, high-quality Chinese multimodal pre-training data, significantly contributing to the development of multimodal models targeting Chinese scenarios.

    \item LAION~\cite{schuhmann2022laion}: The LAION (Large-scale Artificial Intelligence Open Network) project provides one of the largest publicly available image-text pair datasets. For instance, LAION-5B contains nearly 6 billion image-text pairs crawled from the web. This data is filtered using CLIP models to ensure a degree of relevance between images and texts. Due to its immense scale and diversity, the LAION dataset has become fundamental for pre-training many large multimodal models. Its subset, Laion-COCO \cite{schuhmann2022laioncoco}, contains 600 million samples with high-quality captions and aims to provide a large-scale dataset stylistically closer to MS COCO~\cite{lin2014microsoft}.
    
    \item COYO~\cite{byeon2022coyo}: COYO is another large-scale image-text pair dataset, comprising approximately 747 million samples. Similar to LAION, it is sourced from web crawls and undergoes filtering processes. It offers the community an alternative large-scale pre-training resource to LAIONl.
    
    \item DataComp~\cite{gadre2023datacomp}: %DataComp introduces the concept of ``Dataset Computation," aiming to optimize the dataset creation and filtering process competitively, rather than solely focusing on model architecture. Their release, 
    DataComp, contains 1.4 billion samples derived from Common Crawl using carefully designed filtering strategies (CLIP score and Image-based filtering), intended to provide higher quality image-text pairs than raw crawled data.

    \item ShareGPT4V~\cite{chen2024sharegpt4v}: This dataset provides approximately 100K high-quality image-text conversational data points. It is specifically designed and used to enhance the instruction-following and dialogue capabilities of large multimodal models, making them better conversational agents.

    \item ALLaVA~\cite{chen2024allava}: This dataset, comprising 1.4 million samples, is synthetically generated to facilitate the training of resource-friendly Lite Vision-Language Models (LVLMs). The generation pipeline leverages strong proprietary models (like GPT-4V) in a multi-stage process: first, images are selected from sources like LAION and Vision-FLAN; then, fine-grained, detailed captions are generated for these images; finally, complex reasoning visual question-answering pairs are created, emphasizing detailed answers that include evidence and chain-of-thought, to support robust visual instruction fine-tuning.

    \item CapsFusion-120M~\cite{yu2024capsfusion}: It is a large-scale collection of 120M image-text pairs selected from Laion-COCO~\cite{schuhmann2022laioncoco}. The caption is acquired by  integrating the captions in Laion-COCO with CapsFusion-LLaMA~\cite{yu2024capsfusion}.

    \item Cambrian-10M(7M)~\cite{tong2024cambrian}: Cambrian-10M is a large-scale dataset designed for multimodal instruction tuning, sourced from a diverse array of data with an unbalanced distribution across categories. To enhance the quality of the dataset, data filtering based on a refined data ratio is applied, which results in the creation of Cambrian-7M. 

    \item LLaVA-OneVision~\cite{li2024llava}: This visual instruction tuning collection features two main parts: a Single-Image dataset of 3.2 million diverse, categorized samples (QA, OCR, math, etc.), and the OneVision dataset with 1.6 million mixed-modal samples (including video, multi-image, and selected single-image data).

    \item Infinity-MM~\cite{li2024llava}: Infinity-MM is a comprehensive multimodal training dataset with over 40 million samples, created by extensively collecting and categorizing existing open-source datasets alongside newly generated data. This collection includes image captions, general visual instructions, higher-quality selective instructions, and a significant portion of data generated by GPT-4 or synthesized using a custom VLM-based pipeline to ensure alignment and diversity. All data undergoes rigorous processing and filtering for quality and consistency.

    \item Other Datasets:  Additional understanding datasets developed recently include GRIT (Grid-based Representation for Image-Text)~\cite{peng2023kosmos} (20M samples emphasizing fine-grained image region-text phrase alignment). Furthermore, while SAM Dataset~\cite{kirillov2023segment} does not initially consist of image-text pairs, the collection of 11 million high-resolution images with detailed segmentation masks offers valuable spatial and semantic information. It can enhance the fine-grained understanding capabilities of multimodal models, like comprehending object locations, boundaries, or performing region-specific operations. In addition, data for text-to-image models can also be used for multimodal understanding task.

\end{itemize}

\subsection{Text-to-Image Datasets}

These datasets are mainly used for training models that generate images corresponding to textual descriptions. They typically consist of image-text pairs, often with a higher emphasis on the aesthetic quality of the images, the richness of the content, or specific stylistic attributes.

\begin{itemize}
    \item CC-12M (Conceptual Captions 12M) \cite{changpinyo2021conceptual}: CC-12M contains about 12 million image-text pairs extracted and filtered from web Alt-text. Compared to raw web-crawled data, its textual descriptions are generally more concise and descriptive, making it widely used for training text-to-image models.

    \item LAION-Aesthetics \cite{schuhmann2022laion}: This is a subset of the LAION dataset, filtered using an aesthetic scoring model to select approximately 120 million images (and their texts) deemed to have higher ``aesthetic value". %This dataset is crucial for training T2I models like Stable Diffusion to generate visually appealing, high-quality images.

    % \item Mario-10M \cite{chen2024textdiffuser} and AnyWord-3M \cite{tuo2023anytext}: These two datasets focus on accurate text rendering within images. Mario-10M (10M samples), used for training the TextDiffuser model~\cite{chen2024textdiffuser}, and AnyWord-3M (3M samples), used to train AnyText~\cite{tuo2023anytext}, provide data specifically designed to improve the legibility and placement of text generated in images.

    \item Text Rendering Datasets: Several datasets have been developed to specifically address the challenges of accurately and legibly rendering text within generated images. Mario-10M \cite{chen2024textdiffuser}, with 10 million samples, was used to train the TextDiffuser model~\cite{chen2024textdiffuser}, providing data designed to improve text placement and legibility. The RenderedText dataset~\cite{RenderedText} offers 12 million high-resolution synthetic images of handwritten text, generated with diverse visual attributes, serving as a rich resource for handwritten text understanding and generation. AnyWord-3M \cite{tuo2023anytext}, containing 3 million samples, is crucial for training models like AnyText~\cite{tuo2023anytext} and also focuses on enhancing the quality of generated text. Lastly, TextAtlas5M~\cite{wang2025textatlas5m} targets dense text generation, incorporating a diverse mix of interleaved documents, synthetic data, and real-world images with longer captions and human annotations to tackle complex text-rich image scenarios.

    \item JourneyDB \cite{sun2023journeydb}: JourneyDB consists of 4 million high-quality image-prompt pairs generated by the Midjourney platform~\footnote{www.midjourney.com}. As Midjourney is known for generating creative and artistic images, this dataset provides valuable resources for training models to learn complex, detailed, and artistically styled text-to-image mappings.

    \item CosmicMan-HQ 1.0~\cite{li2024cosmicman}: It comprises 6 million high-quality real-world human images with an average resolution of 1488 × 1255 pixels. This dataset is distinguished by its precise text annotations, derived from 115 million attributes varying in granularity. It can be used to improve the capability of generating human images.

    \item DOCCI~\cite{onoe2024docci}: DOCCI provides 15k uniquely curated images, each with long, human-annotated English descriptions (average 136 words) designed to be highly detailed and to differentiate between similar images. The dataset's focus on fine-grained descriptions and contrastive image sets makes it a valuable resource for training and evaluating both image-to-text and text-to-image models, particularly for their ability to handle nuanced details and complex compositions.

    \item PixelProse~\cite{singla2024pixels}: PixelProse extracted from DataComp~\cite{gadre2023datacomp}, CC-12M~\cite{changpinyo2021conceptual}, and RedCaps~\cite{desai2021redcaps}, contains richly annotated images with corresponding textual descriptions. This dataset provides valuable metadata such as watermark presence and aesthetic scores which can be used for filtering to get expected images.

    \item Megalith~\cite{BoerBohan2024Megalith10m}: Megalith is a dataset consisting of approximately 10 million links to Flickr images categorized as ``photo" with licenses ensuring no copyright restrictions. The captions made by the community using models like ShareCaptioner~\cite{chen2024sharegpt4v}, Florence2~\cite{xiao2024florence}, and InternVL2~\cite{chen2024far,chen2024internvl} are available publicly.

    \item PD12M~\cite{meyer2024public}: PD12M consists of 12.4 million high-quality public domain and CC0-licensed images paired with synthetic captions generated using Florence-2-large~\cite{xiao2024florence}. It is designed for training text-to-image models, offering a substantial collection while minimizing copyright concerns. 

    \item Synthesized Datasets: Specialized datasets for text-to-image synthesis are increasingly created using existing generative models. The text-to-image-2M dataset~\cite{text-to-image-2M} provides ~2 million enhanced text-image pairs for fine-tuning, curated using advanced T2I and captioning models. SFHQ-T2I~\cite{david_beniaguev_2024_SFHQ_T2I} offers 122K diverse, high-resolution synthetic face images generated by multiple T2I models, ensuring variance and privacy. For entity control, the EliGen TrainSet~\cite{zhang2025eligen} uses images from a baseline model (FLUX.1-dev) and MLLM-generated prompts for stylistic consistency and detailed annotation. Similarly, BLIP-3o 60k~\cite{chen2025blip3} provides 60,000 instruction tuning samples distilled from GPT-4o, covering various categories for diverse training. ShareGPT-4o-Image\cite{sharegpt4oimage} contributes 45K text-to-image pairs, where prompts are generated through both a structured attribute-first approach and an image-first approach, with corresponding images synthesized by GPT-4o's image generation capabilities to distill its advanced skills. To specifically address blind spots in real-world data, Echo-4o-Image~\cite{ye2025echo} provides over 100K samples targeting surreal fantasy scenarios and complex, long-tail instructions to enhance model imagination and alignment.

    \item Other Datasets: SAM dataset~\cite{kirillov2023segment} (approx. 11 M high-resolution images) and DenseFusion \cite{li2024densefusion} (1M samples) are other potential data sources for text-to-image generation model training. Note that, the multimodal understanding datasets can be utilized for synthesizing text-to-image generation data via aesthetics score filtering, NSFW filtering, resolution filtering, watermark filtering, recaption, etc., which is not introduced here. 
\end{itemize}

\subsection{Image Editing Datasets}

With advancing model capabilities, instruction-based image editing has become an important research direction. Datasets in this category typically contain triplets of (source image, editing instruction, target image). These datasets are utilized to train models to alter input images according to textual commands, thereby enhancing both the comprehension and generation capabilities of unified models.

\begin{itemize}
    \item InstructPix2Pix \cite{brooks2023instructpix2pix}: This dataset was generated using an innovative synthetic approach: first, a large language model (like GPT-3) generates an editing instruction and a caption for the target image; then, a text-to-image model (like Stable Diffusion) generates the ``before" and ``after" images based on the original and target captions. This method automatically created about 313K (instruction, input image, output image) training samples.

    \item MagicBrush \cite{zhang2023magicbrush}: MagicBrush is a high-quality, manually annotated dataset for instruction-based image editing. It contains approximately 10K samples covering various realistic and fine-grained editing operations (like object addition/removal/replacement, attribute modification, style transfer) and provides masks for the edited regions. Its manual annotation leads to more natural and diverse instructions.

    \item HIVE~\cite{zhang2023hive}: The HIVE framework introduces human feedback into instructional visual editing, providing a 1.1M training dataset (generated similarly to InstructPix2Pix using GPT-3 and Prompt-to-Prompt, plus cycle consistency augmentation) and a 3.6K reward dataset where humans rank model outputs. 

    \item EditWorld~\cite{yang2024editworld}: EditWorld introduces the task of ``world-instructed image editing," focusing on realistic world dynamics. Its dataset is curated through two branches: one uses GPT-3.5 for world instructions and T2I models for complex input-output image generation, and the other extracts paired frames from videos with vision-language models generating corresponding instructions for dynamic transformations. 

    \item PromptFix~\cite{zeng2024promptfix}: PromptFix constructs a large-scale instruction-following dataset (~1.01M triplets) focusing on a comprehensive range of image processing tasks, particularly low-level tasks (e.g., inpainting, dehazing, super-resolution, colorization). 

    \item HQ-Edit \cite{hui2024hq}, SEED-Data-Edit \cite{ge2024seed}, UltraEdit \cite{zhao2024ultraedit}, OmniEdit \cite{wei2024omniedit}, AnyEdit \cite{yu2024anyedit}: These represent more recent, larger-scale image editing datasets. For instance, SEED-Data-Edit contains 3.7M samples, UltraEdit has 4M samples, AnyEdit provides 2.5M samples, OmniEdit includes 1.2M samples, and HQ-Edit contains 197K samples. They often combine automated generation with human filtering/annotation, aiming to provide larger-scale, higher-quality, and more diverse editing instructions and image pairs to train more robust instruction-following editing models. 

    \item RefEdit~\cite{pathiraja2025refedit}: This synthetic dataset specifically targets instruction-based editing challenges involving referring expressions in complex scenes. It's generated using GPT-4o for text components (prompts, instructions, referring expressions), FLUX for initial images, Grounded SAM for precise mask generation from expressions, and specialized models for controlled edits like object removal or modification. 

    \item ImgEdit~\cite{ye2025imgedit}: ImgEdit is a large-scale (1.2M edit pairs) dataset designed for high-quality single-turn and multi-turn image editing. Its multi-stage generation pipeline filters LAION-Aesthetics, uses vision-language models and detection/segmentation models for grounding and instruction generation (including spatial cues and multi-turn dialogues), employs state-of-the-art generative models (FLUX, SDXL with plugins) for task-specific inpainting, and uses GPT-4o for final quality filtering.

    \item ByteMorph-6M~\cite{chang2025bytemorph}: ByteMorph-6M is a large-scale dataset with over 6 million image editing pairs specifically designed for instruction-guided editing involving non-rigid motions (e.g., camera viewpoint shifts, object deformations, human articulations). It is constructed by first using a Vision-Language Model to generate ``Motion Captions" from instruction templates for an initial frame; then, an image-to-video model generates a dynamic video based on this motion caption; finally, frames are sampled from these videos, and an LLM generates precise editing instructions describing the transformation between neighboring frame pairs, which form the source-target editing data.

    \item ShareGPT-4o-Image (Editing)~\cite{sharegpt4oimage}: Complementing its text-to-image data, ShareGPT-4o-Image also includes 46K instruction-guided image editing triplets. These samples are generated by first selecting a source image (either from its text-to-image collection or real photos), then sampling an editing task from a predefined taxonomy (e.g., object manipulation, style transfer), having an LLM synthesize a natural language instruction for that task and image, and finally using GPT-4o's image generation capabilities to produce the edited image.

    \item GPT-Image-Edit-1.5M~\cite{wang2025gpt}: GPT-IMAGE-EDIT-1.5M is a large-scale image editing dataset containing over 1.5 million high-quality instruction-guided image editing triplets. It is constructed by leveraging the powerful capabilities of GPT-4o to systematically unify and refine three existing datasets: OmniEdit, HQ-Edit, and UltraEdit. The core methodology involves regenerating output images to enhance visual quality and instruction alignment, as well as selectively rewriting prompts to improve semantic clarity. This process results in a high-fidelity corpus designed to bridge the gap between proprietary and open-source instruction-guided image editing models.

    \item X2Edit~\cite{ma2025x2edit}: X2Edit is a large-scale and comprehensive image editing dataset with 3.7 million samples, designed to be balanced across 14 diverse editing tasks. It is constructed through an automated pipeline that first uses a VLM to generate task-aware instructions, which are then executed by industry-leading and expert generative models to produce the edited images. To overcome the quality and balance issues in existing open-source resources, all generated pairs undergo a final, rigorous filtering stage based on multiple scoring mechanisms to ensure high fidelity and accuracy.

\end{itemize}

\subsection{Interleaved Image-Text Datasets}
Beyond datasets consisting of paired images and captions, another important category comprises interleaved image-text data. These datasets contain documents or sequences where text and images naturally follow one another, mirroring content found on webpages or in documents. Training models on this interleaved data enhances their capability to comprehend and generate multimodal content, an essential goal for unified models.

\begin{itemize}

\item Multimodal C4 (MMC4)~\cite{zhu2023multimodal}: MMC4 augments the large-scale text-only C4~\cite{raffel2020exploring} corpus by algorithmically interleaving images into the text documents sourced from Common Crawl. This public dataset, containing over 101 million documents and 571 million images, was created to provide the necessary interleaved pre-training data for models designed to process mixed sequences of images and text. %Its release supports research into multimodal few-shot learning and understanding complex interactions between multiple images within a document context.

\item OBELICS~\cite{laurenccon2023obelics}: OBELICS is an open, web-scale dataset comprising 141 million multimodal web documents extracted from Common Crawl, featuring 353 million images interleaved with 115 billion text tokens. 
%It was developed to provide a publicly available and well-documented resource for training large models on naturally occurring interleaved data, addressing the lack of access to similar datasets used previously. 
The dataset focuses on capturing the full document structure rather than isolated image-text pairs, aiming to improve model performance on various benchmarks.
% as demonstrated by the IDEFICS models trained on it.

\item CoMM~\cite{chen2024comm}: CoMM is a high-quality, curated dataset focused specifically on the coherence and consistency of interleaved image-text sequences, containing approximately 227K samples. It addresses limitations in narrative flow and visual consistency observed in larger datasets by sourcing content primarily from instructional and visual storytelling websites (like WikiHow) and applying a multi-perspective filtering strategy. CoMM aims to enhance MLLMs' ability to generate logically structured and visually consistent multimodal content and introduces new benchmark tasks specifically designed to evaluate these capabilities.

\item OmniCorpus~\cite{li2024omnicorpus}: OmniCorpus is a very large-scale (10 billion-level) image-text interleaved dataset, containing 8.6 billion images and 1,696 billion text tokens from 2.2 billion documents. It was created using an efficient data engine that extracts and filters content from diverse sources, including English and non-English websites as well as video platforms (extracting keyframes and transcribing audio). The dataset incorporates human-feedback filtering to enhance data quality, aiming to provide a solid foundation for MLLM research.

\end{itemize}

\subsection{Other Text+Image-to-Image Datasets}

Beyond the previously mentioned categories, to further enhance a unified model's capabilities—such as generating images based on provided subject images, or utilizing control signals (e.g., depth maps, canny maps)
%, or creating new perspectives for an image
—we introduce relevant datasets in this section.

\begin{itemize}
    \item LAION-Face~\cite{zheng2022general}: The datasets discussed above emphasize general subject-driven generation, whereas ID-preserving image generation represents a specialized subset of this category. Utilizing LAION-Face, which includes 50 million image-text pairs, recent advancements such as InstantID~\cite{wang2024instantid} have succeeded in generating images while maintaining character identity.

    \item MultiGen-20M~\cite{qin2023unicontrol}: This dataset comprises 20 million samples designed to train models capable of unified image generation conditioned on multiple control signals (e.g., text descriptions, edge maps, depth maps, segmentation masks, sketches), such as UniControl~\cite{qin2023unicontrol}. It integrates data from various sources and converts them into a unified format, enabling models to learn multi-task, multi-conditional image generation. The dataset can be structured as triples, such as ``depth map, instruction with prompt, target image" (The ``instruction with prompt"  might be phrased as: ``Generate an impressive scene following the depth map."), to effectively train unified models.

    \item Subjects200K~\cite{tan2024ominicontrol}: Containing 200K samples, Subjects200K focuses on subject-driven image generation, crucial for personalized content creation. This dataset was generated synthetically through a multi-stage pipeline: initially, an LLM (ChatGPT-4o) creates structured descriptions involving object categories and scenes; subsequently, an image synthesis model (FLUX~\cite{flux2024}) generates diverse yet consistent paired images based on these descriptions; finally, the LLM performs quality assessment on the generated pairs to ensure subject consistency, proper composition, and high resolution.

    \item SynCD~\cite{kumari2025syncd}: SynCD (Synthetic Customization Dataset) provides approximately 95K sets of images specifically designed for text+image-to-image customization tasks, addressing the lack of public datasets containing multiple images of the same object under diverse conditions. It is synthesized by leveraging existing text-to-image models and 3D asset datasets (like Objaverse~\cite{deitke2023objaverse}) to generate multiple consistent views of an object with varied lighting, backgrounds, and poses, incorporating techniques like shared attention and depth guidance. 

    \item X2I-subject-driven~\cite{xiao2024omnigen}: The X2I-subject-driven dataset facilitates subject-driven image generation through two components. The GRIT-Entity dataset is derived from the GRIT~\cite{peng2023kosmos} dataset by automatically detecting and segmenting objects from images, followed by an optional MS-Diffusion~\cite{wang2024ms} repainting step to improve quality and diversity. To encourage more robust generative capabilities beyond simple copy-paste patterns, the higher-quality Web-Images dataset was constructed by identifying notable individuals through automated text analysis and large language model filtering, scraping their web images, performing automated visual verification to ensure subject accuracy, and then captioning the selected images.

    \item Graph200K~\cite{li2025visualcloze}: Graph200K is a graph-structured dataset built upon Subjects200K, where each image is augmented with 49 types of annotations spanning five meta-tasks: conditional generation (e.g., canny edges, depth maps, segmentation), IP preservation, style transfer (semantic-variant and -invariant), image editing (background-variant and -invariant using VLMs and inpainting models), and restoration (via online degradations). This structure aims to increase task density and inter-relation, enabling models to learn shared and transferable knowledge for universal image generation by formulating tasks as paths within this graph.

    \item Echo-4o-Image (Multi-Reference)~\cite{ye2025echo}: This component of the dataset addresses the scarcity of structured, multi-input generation tasks in natural image collections. It provides 73K synthetic samples for ``Multi-to-one" generation, which are explicitly designed with diverse instructions and rich reference information. This controlled synthesis offers a more targeted and varied training source for multi-reference image composition compared to alternatives like sampling from video frames.
   
   % \item BlendedMVS~\cite{yao2020blendedmvs},
   %MVS-Synth~\cite{huang2018deepmvs},
   % ScanNet~\cite{dai2017scannet},
   % Replica~\cite{straub2019replica},
   % MVImageNet~\cite{yu2023mvimgnet}, and
   % Objaverse~\cite{deitke2023objaverse} are datasets from the 3D computer vision community. These datasets can be utilized to create data for generating new perspectives of a given image and producing multi-view images for a given subject, thereby enhancing the capabilities of unified models.

\end{itemize}

Subject-driven generation, involving both single and multiple subjects, is a crucial image generation capability that is increasingly attracting attention within the community. It is also anticipated to be a significant feature inherent in unified models. However, obtaining such specialized data from public datasets is challenging, leading to the frequent use of data synthesis methods, exemplified by datasets like Subjects200K and SynCD. These datasets illustrate the growing reliance on synthetic data to address the shortage of publicly available training examples needed for tasks like subject-driven generation and customization.

To create large-scale datasets, various pipelines~\cite{wang2024ms,pan2023kosmos,chen2024unireal,huang2025wegen} have been developed to programmatically generate suitable training data, typically utilizing readily accessible image or video sources. Below, we provide a brief overview of these pipelines for reference.

\begin{itemize}
    
    \item Data synthesis from images: These pipelines often start with single images, using models like BLIP-2~\cite{li2023blip} or Kosmos2~\cite{peng2023kosmos} for initial captioning (including grounding captions with bounding boxes), followed by object detection (e.g., Grounding DINO~\cite{liu2024grounding}) and segmentation (e.g., SAM~\cite{kirillov2023segment}) to extract subject masks and region captions. These pipelines can generate data for single subject customization and multiple subjects customization.
    
    \item Data synthesis from videos: 
    Data constructed from images often cause the copy-paste issue in model learning. The pipeline of synthesizing data from videos can alleviate this issue by extracting subjects from different frames with video segmentation models (e.g., SAM2~\cite{ravi2024sam2}). In addition, this pipeline can also enable the generation of training data for image editing task.~\cite{chen2024unireal}. 
    %These pipelines generate video-level captions and often extract key frames (e.g., two frames) to serve as before/after pairs for editing tasks. LLMs like GPT-4o can generate editing instructions based on frame differences, while video segmentation models (e.g., SAM2~\cite{ravi2024sam2}) provide tracked masks. This data is valuable for instruction-based image editing, and can also contribute to subject-driven tasks by extracting consistent subjects across frames~\cite{wang2024ms}. 
    % MLLMs often play a role in generating instructions or filtering the synthesized data for quality.
    
\end{itemize}

Robust unified multimodal models rely critically on large-scale, high-quality, and diverse training datasets developed recently, encompassing image-text pairs, interleaved image-text documents, and task-specific formats. While massive web-scale paired data (like LAION, COYO) and interleaved document corpora (like MMC4, OBELICS) provide broad semantic coverage and contextual understanding for pre-training, significant efforts focus on enhancing data quality and tailoring resources for specific attributes or advanced capabilities. Specialized datasets are increasingly crucial for improving instruction-based editing, accurate text rendering, coherent multimodal generation, and complex conditional control. Furthermore, recognizing the scarcity of high-quality public data for tasks like instruction-based image editing and subject customization, the development and utilization of data synthesis pipelines have become essential, enabling the creation of targeted datasets needed to train these highly specific model functionalities. Ultimately, the continuous evolution, growing scale, targeted specialization, and innovative synthesis of these varied data resources are the fundamental drivers enabling the increasingly sophisticated understanding and generation capabilities of unified multimodal models.

%% file: sec/5_benckmarks.tex
\begin{table*}
\centering
\footnotesize  
\caption{Statistical summary of current evaluations and benchmarks for unified large-scale generative models. This table categorizes benchmarks into \textit{Understanding, Image Generation, and Interleaved Generation}, detailing the size, description, input/output types, and publication venues for each.}
\begin{tabular}{c|c|c|c|c}
    % \toprule[1pt]
    \hline
    Benchmark & Size & Description & In.out Type & Venue \\
   \hline
   \multicolumn{5}{l}{\emph{Understanding}} \\
   \hline
   VQA~\cite{antol2015vqa} & 10M QAs &	Open-domain Visual QA & Image + Question $\rightarrow$ Answer & ICCV2015 \\
    VQAv2~\cite{goyal2017making} & 1M QAs & Open-domain Visual QA & Image + Question $\rightarrow$ Answer & CVPR2017 \\
   CLEVR~\cite{clevr17} & 853K QAs & Compositional Visual QA &	Image + Question $\rightarrow$ Answer & CVPR2017 \\
   GQA~\cite{gqa19} & 22M QAs & Compositional Visual QA & Image + Question $\rightarrow$ Answer & CVPR2019 \\
   OK-VQA~\cite{okvqa19} & 14K QAs & Knowledge-based VQA & Image + Question $\rightarrow$ Answer & CVPR2019 \\
   VCR~\cite{vcr19} & 290K QAs & Commonsense Visual QA & Img. + Q. $\rightarrow$ Answer + Rationale & CVPR2019 \\
   VisDial~\cite{das2017visual} & 1.2M Dialogs & Multi-turn Visual Dialog & Image + Dialog $\rightarrow$ Answer & CVPR2019 \\
   ChartQA~\cite{masry2022chartqa} & 32.7K QAs & Data Visualization QA & Image + Question $\rightarrow$ Answer & ACL2020 \\
   TextVQA~\cite{singh2019towards} & 45K QAs & Scene Text Visual QA & Image + Question $\rightarrow$ Answer & CVPR2020 \\
   A-OKVQA~\cite{aokvqa22} & 25K QAs & Expanded Commonsense VQA & Image + Question $\rightarrow$ Answer & ECCV2022 \\	
   HaluEval~\cite{li2023halueval} & 35K Samples & Hallucination Detection & Model output $\rightarrow$ Yes / No & EMNLP2023 \\
   VSR~\cite{liu2023visual} & 3K QAs & Spatial Reasoning & Image + Question $\rightarrow$ True / False & TACL2023 \\	
   LAMM~\cite{yin2023lamm} & 62K QAs & Instruction Benchmarking & Features + Instruction $\rightarrow$ Output & NeurIPS2023 \\
   LLaVa-Bench~\cite{visual_instruction_tuning} & 150 QAs & Instruction Benchmarking & Image + Question $\rightarrow$ Answer & NeurIPS2023 \\
   OwlEval~\cite{ye2023mplug} & 82 Qs & Visual-related Eval & Image + Instruction $\rightarrow$ Answer & Arxiv2023 \\
   MMBench~\cite{liu2024mmbench} & 3K QAs & Fine-grained Multi-modal Eval & Image + Question $\rightarrow$ Answer & ECCV2024 \\
   MMMU~\cite{yue2024mmmu} & 11.5K QAs & Expert-level Understanding &	Image + Question $\rightarrow$ Answer & CVPR2024 \\
   MM-Vet~\cite{yu2023mm} & 218 Samples & VL Capability Eval & Image + Question $\rightarrow$ Answer & ICML2024 \\
   MM-Vet v2~\cite{yu2024mm} & 218+ Samples & VL Sequence Understanding &Image + Sequences $\rightarrow$ Answer & Arxiv2024 \\
   MMStar~\cite{chen2024we} & 1.5K QAs & Vision Indispensable Eval & Image + Question $\rightarrow$ Answer & NeurIPS2024 \\
   SEED-Bench~\cite{li2023seed} & 19K QAs & Comprehensive Evaluation & Image/Video + MCQ $\rightarrow$ Answer & CVPR2024 \\
   Open-VQA~\cite{ging2024open} & Varied & VQA Evaluation & Image + Q/A $\rightarrow$ QA Chain & ICLR2024 \\
   MathVista~\cite{mathvista24} & 6K QAs & Math Reasoning & Image + Text $\rightarrow$ Math Output & ICLR2024 \\
   General-Bench~\cite{generalbench} & $>$700 tasks & Ultra Large-scale Eval & Varied by Task & Arxiv2025 \\
   \hline
   \multicolumn{5}{l}{\emph{Image Generation}} \\
   \hline
   DrawBench~\cite{saharia2022photorealistic} & 200 Prompts & Comprehensive Eval & Text Prompt $\rightarrow$ Image & NeurIPS2022 \\
   PartiPrompts~\cite{partiprompts} & 1600 Prompts & Comprehensive Eval & Text Prompt $\rightarrow$ Image & TMLR2022 \\ 
   PaintSkills~\cite{dalleval} & $\sim$7K Scenes & Compositional Eval & Text Prompt $\rightarrow$ Image & ICCV2023 \\
   HRS-Bench~\cite{bakr2023hrs} & 960 Prompts & Multi-skill Eval & Text Prompt $\rightarrow$ Image & ICCV2023 \\
   TIFA~\cite{hu2023tifa} & 4081 Prompts & QA-based Eval & Text Prompt $\rightarrow$ Image & ICCV2023 \\
   GenEval~\cite{ghosh2023geneval} & 1000 Prompts & Object-focused Eval  & Text Prompt $\rightarrow$ Image & NeurIPS2023 \\
   T2I-CompBench~\cite{huang2023t2i} & 6000 Prompts & Compositional Eval & Text Prompt $\rightarrow$ Image & NeurIPS2023 \\	
   HEIM~\cite{lee2023holistic} & $\sim$1620 Prompts & Comprehensive Eval & Text Prompt $\rightarrow$ Image & NeurIPS2023 \\	
   Commonsense-T2I~\cite{fu2024commonsense} & 500 Prompts & Commonsense-driven Eval & Text Prompt $\rightarrow$ Image & COLM2024 \\
   DSG-1k~\cite{DSG} & 1060 Prompts & Compositional Eval & Text Prompt $\rightarrow$ Image & ICLR2024 \\
   GenAI-Bench~\cite{li2024genai} & 1600 Prompts & Compositional Eval & Text Prompt $\rightarrow$ Image & CVPR2024 \\
   ConceptMix~\cite{conceptmix} & 2100 Prompts &	Compositional Eval & Text Prompt $\rightarrow$ Image & NeurIPS2024 \\
   DPG-Bench~\cite{dpg_bench} & 1065 prompts & Attribute Eval & Text Prompt $\rightarrow$ Image & Arxiv2024\\
   T2I-CompBench++~\cite{huang2025t2i} & 6000+ Prompts & Compositional Eval & Text Prompt $\rightarrow$ Image & TPAMI2025 \\
   MMIG-Bench~\cite{hua2025mmig} & 4850 Prompts & Comprehensive Eval & Text Prompt $\rightarrow$ Image & Arxiv2025 \\
   OneIG-Bench~\cite{oneig} & $\sim$2k Prompts & Comprehensive Eval & Text Prompt $\rightarrow$ Image & Arxiv2025 \\
   WISE~\cite{niu2025wise} & 1k Prompts & World Knowledge Eval & Text Prompt $\rightarrow$ Image & Arxiv2025 \\
   CVTG-2K~\cite{cvtg2k} & 2k Prompts & Multi-region Visual Text Eval & Text Prompt $\rightarrow$ Image & Arxiv2025 \\
   WorldGenBench~\cite{WorldGenBench} & 1072 Prompts & World Knowledge Eval & Text Prompt $\rightarrow$ Image & Arxiv2025 \\

   EditBench~\cite{editbench} & 240 Edits & Mask-guided Editing & Img. + Ins. + [Mask] $\rightarrow$ Image & CVPR2023 \\	
   MagicBrush~\cite{zhang2023magicbrush} & 1053 Edits & Real‑image Editing & Image + Instruction $\rightarrow$ Image & NeurIPS2023 \\
   EditVal~\cite{basu2023editval} & 648 Edits & Attribute‑focused Eval & Image + Instruction $\rightarrow$ Image & Arxiv2023 \\
   Emu-Edit~\cite{sheynin2024emu} & 3055 Edits & Multi‑task Editing & Image + Instruction $\rightarrow$ Image & CVPR2024 \\
   Reason-Edit~\cite{reasonedit} & 219 Edits & Complex Instruction Editing & Image + Instruction $\rightarrow$ Image & CVPR2024 \\
   I2EBench~\cite{ma2024i2ebench} & 2240 Edits &	Multi‑dimensional Eval & Image + Instruction $\rightarrow$ Image & NeurIPS2024 \\
   HumanEdit~\cite{bai2024humanedit} & 5.7K Edits & Human‑rewarded Editing & Img. + Ins. + [Mask] $\rightarrow$ Image & Arxiv2024 \\	
   HQ-Edit~\cite{hui2024hq} & $\sim$200K Edits & High‑resolution Editing & Image + Instruction $\rightarrow$ Image & ICLR2025 \\
   {AnyEdit~\cite{yu2024anyedit}} & 1250 Edits & Comprehensive Eval & Image + Instruction $\rightarrow$ Image & CVPR2025 \\
   IE-Bench~\cite{iebench} & 301 Edits & Human-aligned Perceptual Eval & Image + Instruction $\rightarrow$ Image & Arxiv2025 \\
   GEdit-Bench~\cite{liu2025step1x} & 606 Edits & Real-world-grounded Editing & Image + Instruction $\rightarrow$ Image & Arxiv2025 \\
   CompBench~\cite{compbench} & 3K Edits & Complex Instruction Editing & Image + Instruction $\rightarrow$ Image & Arxiv2025 \\
   GIE-Bench~\cite{giebench} & 1080 Edits & Content-preserving Eval & Image + Instruction $\rightarrow$ Image & Arxiv2025 \\
    {EditInspector~\cite{yosef2025editinspector}} & 783 Edits & Comprehensive Eval & Image + Instruction $\rightarrow$ Image & Arxiv2025 \\
   {ComplexBench-Edit~\cite{wang2025complexbench}} & $<$1K List of Edits & Chain-dependent Editing Eval & Image + Instruction $\rightarrow$ Image & Arxiv2025 \\
    {ByteMorph-Bench~\cite{chang2025bytemorph}} & 613 Edits & Non-rigid Editing Eval & Image + Instruction $\rightarrow$ Image & Arxiv2025 \\
    {RefEdit-Bench~\cite{pathiraja2025refedit}} & 200 Edits & Expression-driven Editing Eval & Image + Instruction $\rightarrow$ Image & Arxiv2025 \\
    {ImgEdit-Bench~\cite{ye2025imgedit}} & 200 Edits & Expression-driven Editing Eval & Image + Instruction $\rightarrow$ Image & Arxiv2025 \\
    {KRIS-Bench~\cite{wu2025kris}} & 1267 Edits & Cognitive Reasoning Eval & Image + Instruction $\rightarrow$ Image & Arxiv2025 \\
   \hline
   \multicolumn{5}{l}{\emph{Interleaved / Compositional Generation}} \\
   \hline
   InterleavedBench~\cite{liu2024holistic} & 815 Samples & Human-curated Interleaving & Text + Images $\rightarrow$ Text + Images & EMNLP2024 \\
   OpenLEAF~\cite{an2023openleaf} & 30 Queries & Open-domain Interleaving & Query $\rightarrow$ Text + Images	& MM2024 \\
   ISG~\cite{chen2024interleaved} & 1150 Samples & Scene‑driven Interleaving &	Graph + Text → Text + Images & ICLR2025 \\
   MMIE~\cite{xia2024mmie} & 20K Queries & Knowledge‑intensive Interleaving & History + Query $\rightarrow$ Response & ICLR2025 \\
   OpenING~\cite{zhou2024gate} & 5.4K Samples & Open‑domain Interleaving &	Query → Text + Images & CVPR2025 \\	
   UniBench~\cite{unieval} & 81 fine-grained tags & Unified Compositional Eval &	Prompt → Images + Answer & Arxiv2025 \\	
   \hline
   \multicolumn{5}{l}{\emph{Other Types}} \\
   \hline
   MultiGen-20M~\cite{qin2023unicontrol} & Varied & Controllable Generation & Featues + Instruction $\rightarrow$ Image & NeurIPS2023 \\
   Dreambench~\cite{ruiz2023dreambooth} & 30 objects & Subject-Driven Generation & Ref Img. + Instruction $\rightarrow$ Image & CVPR2023 \\
   Dreambench++~\cite{peng2024dreambench++} & 150 imgs & Personalized Generation & Ref Img. + Instruction $\rightarrow$ Image & ICLR2025 \\
   VTBench~\cite{vtbench} & Varied & Visual Tokenizer Eval & Image $\rightarrow$ Reconstructed Image & Arxiv2025 \\
   % \bottomrule[1pt]
   \hline
\end{tabular}
\label{table:benchmark}
\end{table*}

\section{Benchmarks}

Modern large-scale unified multimodal models should not only align visual and linguistic information at the pixel level but also perform complex reasoning, support coherent multi-turn dialogue and integrate external knowledge. Simultaneously, these models are expected to produce high-fidelity visual outputs that faithfully adhere to textual prompts while providing users with fine-grained control over stylistic and compositional elements. In this section we systematically summarize the related evaluation benchmarks. Please refer to Tab.~\ref{table:benchmark} for statistical summary.

\subsection{Evaluation on Understanding}

\noindent\textbf{Perception.} Modern vision‐language large models must accurately connect visual inputs with linguistic descriptions through grounding, recognition and retrieval. Early image–text retrieval and captioning benchmarks such as Flickr30k \cite{flickr30k}, MS COCO Captions \cite{chen2015microsoft} evaluate whether models can retrieve relevant captions and localize textual phrases to image regions. Visual question answering benchmarks like VQA \cite{antol2015vqa}, VQA v2 \cite{goyal2017making}, VisDial \cite{das2017visual} and TextVQA \cite{singh2019towards} further require models to interpret complex scenes and answer free‐form queries about objects, attributes and relationships. Domain‐specific challenges such as ChartQA~\cite{masry2022chartqa} assess understanding of structured charts and graphs, while VSR \cite{liu2023visual} probes spatial relation reasoning in real‐world images.

To unify the evaluation, large‐scale meta‐benchmark suites test both low‐level perception and expert reasoning. MMBench~\cite{liu2024mmbench} supplies 3K bilingual multiple‐choice questions spanning grounding, recognition and retrieval, enabling cross‐lingual comparison. MMMU~\cite{yue2024mmmu} adds about 11.5K college‐level multimodal problems across six disciplines to probe domain knowledge and logical deduction. HaluEval~\cite{li2023halueval} diagnoses hallucination recognition on a diverse set of model-generated and annotated statements. MM-Vet~\cite{yu2023mm} covers recognition, OCR, spatial reasoning, maths and open question answering, and its v2~\cite{yu2024mm} further evaluates interleaved image–text sequences. SEED-Bench~\cite{li2023seed} designs a pipeline for generating multiple-choice questions that target specific evaluation dimensions and finally offers 19K multi-choice items over 12 dimensions. LLaVa-Bench~\cite{visual_instruction_tuning} provides COCO~\cite{lin2014microsoft} and in-the-wild image sets with dense queries for generalization checks. LAMM~\cite{yin2023lamm} supplies instruction-tuning examples covering 2D and 3D modalities for agent development. Open-VQA~\cite{ging2024open} formulates hierarchical follow-up questions to refine coarse VQA answers. OwlEval~\cite{ye2023mplug} offers human-rated open-ended visual questions assessing relevance and informativeness. MMStar~\cite{chen2024we} curates carefully balanced challenge samples spanning six core skills and 18 axes for high-precision evaluation.

\noindent\textbf{Reasoning.} Building on perception-level evaluation, reasoning benchmarks probe progressively richer cognitive skills. CLEVR~\cite{clevr17} systematically varies object attributes and spatial relations, forcing models to execute multi-hop programs that test counting, comparison and relational logic. Moving to natural images, GQA~\cite{gqa19} leverages dense scene graphs to generate compositional questions whose functional programs are used to test consistency, grounding and plausibility.  

Commonsense extensions such as OK-VQA~\cite{okvqa19} and its larger successor A-OKVQA~\cite{aokvqa22} select questions whose answers lie outside the image, requiring retrieval or inference over world knowledge bases. VCR~\cite{vcr19} further demands that a model not only choose the correct answer but also justify it by selecting a coherent rationale, thereby coupling recognition with explanation and testing multi-step commonsense chains.

Domain-specific reasoning datasets extend this progression beyond everyday scenes.  ChartQA~\cite{masry2022chartqa} introduces questions that intertwine visual perception with quantitative reasoning over bar, line and pie charts, integrating data extraction, logical comparison and arithmetic calculation.  MathVista~\cite{mathvista24} broadens the scope to mathematical problem solving in visually grounded contexts and combines fine-grained visual understanding with symbolic manipulation across diversified examples. These benchmarks form a layered spectrum that spans structured logical inference, open-domain commonsense, visual explanation and numerically intensive tasks, offering a comprehensive stress-test for multimodal reasoning systems. 

Moreover, General-Bench~\cite{generalbench}, an ultra-large benchmark comprising over 700 tasks and 325,800 instances across varied modalities and capabilities, provides a synergy-driven evaluation suite for multimodal generalist models.

\subsection{Evaluation on Image Generation }

\noindent{\textbf{Text‐to‐Image Generation}.} Early automated metrics such as FID~\cite{heusel2017gans} and CLIPScore~\cite{radford2021learning} established the foundation for evaluating image quality. More recent benchmarks, however, emphasize compositional reasoning, prompt alignment, and real-world applicability. PaintSkills~\cite{dalleval}, DrawBench~\cite{saharia2022photorealistic}, and PartiPrompts~\cite{partiprompts} evaluate core compositional capabilities. GenEval~\cite{ghosh2023geneval} evaluates six fine-grained tasks, including single-object generation, object co-occurrence, counting, color control, relative positioning, and attribute binding by comparing outputs from pretrained detectors against ground-truth annotations. 

Expanding on this, GenAI-Bench~\cite{li2024genai} presents 1.6K meticulously crafted human prompts that cover relational, logical, and attribute-based categories. Its evaluation framework combines human preference judgments with automated alignment scores to provide a comprehensive assessment. In addition, HRS-Bench~\cite{bakr2023hrs} evaluates 13 distinct skills that are grouped into five major categories: accuracy, robustness, generalization, fairness, and bias, thereby ensuring scalable and reliable performance measurement. Moreover, DPG-Bench~\cite{dpg_bench} focuses on dense prompts that describe multiple objects, with each object characterized by a variety of attributes and relationships.

The T2I-CompBench~\cite{huang2023t2i} and its successor T2I-CompBench++~\cite{huang2025t2i} specifically target compositional generalization, testing the generation of novel attribute and relation combinations using detector-based scoring. VISOR~\cite{gokhale2022benchmarking} proposes an automatic method for evaluating the spatial understanding capabilities of generative models. Complementing these, Commonsense-T2I~\cite{fu2024commonsense} challenges models to depict everyday concepts that require commonsense grounding.  

To support large-scale concept diversity, EvalMuse-40K~\cite{han2024evalmuse} provides 40K crowdsourced prompts focusing on nuanced concept representation, and HEIM~\cite{lee2023holistic} identifies 12 aspects, including text-image alignment, image quality, aesthetics, originality, reasoning, knowledge, bias, toxicity, fairness, robustness, multilinguality and efficiency. Considering practical needs, FlashEval~\cite{zhao2024flasheval} shrinks the large-scale evaluation set into diverse smaller ones through iterative search to accelerate the benchmark testing. MEMO-Bench~\cite{zhou2024memo} introduces a comprehensive benchmark for evaluating the emotional understanding and expression capabilities of T2I models and MLLMs. ConceptMix~\cite{conceptmix} evaluates text-to-image models’ compositional generation ability by sampling k-tuples of visual concepts to construct prompts and automatically verifying concept presence in the resulting images using a strong visual language model. TIFA~\cite{hu2023tifa} offers a fine-grained benchmark for evaluating text-to-image faithfulness via visual question answering generated from prompts. 

To enrich dependencies in question generation for VQA-based evaluation of image–prompt alignment, DSG-1k~\cite{DSG} refines its questions using a multi-level semantic graph. MMIG-Bench~\cite{hua2025mmig} introduced a multi-dimensional assessment framework that rigorously examines text-to-image generation models. OneIG-Bench~\cite{oneig} introduces a comprehensive fine-grained evaluation framework for text-to-image models across more dimensions. ~\cite{niu2025wise,WorldGenBench} evaluate text-to-image models’ world knowledge understanding, which emphasizes semantic consistency, realism, and aesthetics. CVTG-2K~\cite{cvtg2k} evaluates visual-text generation on complex multi-region layouts, diverse text attributes, and fine-grained positioning.

\noindent{\textbf{Image Editing}.} Benchmarks for instruction‐guided image editing have grown in scale and scope. MagicBrush~\cite{zhang2023magicbrush} is a large-scale, manually annotated dataset for instruction-guided real image editing that covers diverse scenarios: single-turn, multi-turn, mask-provided, and
mask-free editing. HQ-Edit~\cite{hui2024hq} contains approximately 200K high‐resolution edits with computed alignment and coherence scores, allowing quantitatively assessing the quality of image edit pairs using GPT-4V. 

Building on this, I2EBench~\cite{ma2024i2ebench} consolidates over 2K images and 4K multi-step instructions across 16 editing dimensions. EditVal~\cite{basu2023editval} offers a standardized benchmark with fine-grained edit annotations and an automated evaluation pipeline aligned with human judgment. Emu-Edit~\cite{sheynin2024emu} covers seven editing tasks including background changes, object-level edits, and style modifications, with paired instructions and I/O descriptions. Reason-Edit~\cite{reasonedit} is a diagnostic benchmark targeting causal and counterfactual reasoning, emphasizing object relations, attribute dependencies, and multi-step inference. 

Offering masked input–reference pairs across varied objects, attributes, and scenes, EditBench~\cite{editbench} delivers a diagnostic benchmark for text-guided image inpainting that enables precise evaluation of editing quality. HumanEdit~\cite{bai2024humanedit} includes 5,751 high-resolution images and open-form instructions spanning six edit types, with annotated masks and multi-stage human feedback. IE-Bench~\cite{iebench} provides a human-aligned benchmark for evaluating text-driven image editing quality with diverse edits and perceptual scores.

More recent benchmarks include GEdit-Bench~\cite{liu2025step1x} which features 606 real-world instruction–image pairs, CompBench~\cite{compbench} that decomposes edits into location, appearance, dynamics and object dimensions via large-scale MLLM–and–human collaboration, and GIE-Bench~\cite{giebench} which uses multiple-choice VQA and object-aware masking on over 1,000 examples to evaluate editing accuracy and content preservation. Following this trend, benchmarks like~\cite{wang2025complexbench,yosef2025editinspector,yu2024anyedit,ye2025imgedit} also undertake comprehensive evaluation of text-guided image editing, assessing vision consistency, artifact detection, instruction adherence, visual quality, and detail preservation.

Other benchmarks include ByteMorph-Bench~\cite{chang2025bytemorph} which tackles non-rigid image manipulation, RefEdit-Bench~\cite{pathiraja2025refedit} which evaluates referring-expression–based edits in complex multi-entity scenes, and KRIS-Bench~\cite{wu2025kris} which offers a cognitively grounded suite assessing factual, conceptual and procedural reasoning.

\noindent{\textbf{Other Types of Image Generation}.}
Beyond text-to-image generation and editing, additional benchmarks target conditional and personalized synthesis. MultiGen-20M~\cite{qin2023unicontrol} provides over 20 million image–prompt–condition triplets from LAION-Aesthetics-V2~\cite{schuhmann2022laionAestheticsV2}, supporting automated evaluation across diverse visual conditions. DreamBench~\cite{ruiz2023dreambooth} benchmarks personalized generation using 30 reference objects with curated prompts and human fidelity ratings. DreamBench++~\cite{peng2024dreambench++} scales this to 150 subjects and 1,350 prompts, using advanced vision–language models for human-aligned scoring of concept preservation, composition, and style. Together, these datasets span large-scale automated and fine-grained human-centric evaluation of conditional generation.

VTBench~\cite{vtbench} provides a systematic benchmark for evaluating visual tokenizers in autoregressive image generation across image reconstruction, detail preservation, and text preservation.

\subsection{Evaluation on Interleaved Generation}

Interleaved evaluation benchmarks challenge models to seamlessly alternate between text and image modalities across multiple turns, reflecting realistic dialogue and storytelling scenarios. InterleavedBench~\cite{liu2024holistic} is a representative benchmark carefully curated for the evaluation of interleaved textand-image generation, featuring a rich array of tasks to cover diverse real-world use cases and evaluating models on text quality, perceptual fidelity, multimodal coherence and helpfulness. Building on this, ISG~\cite{chen2024interleaved} introduces scene‐graph annotations and a four‐tier evaluation (holistic, structural, block‐level and image‐specific) over 1K samples in eight scenarios and 21 subtasks, enabling fine‐grained assessment of interleaved text–image outputs.

Other benchmarks emphasize open‐domain instruction and end‐to‐end interleaving. OpenING~\cite{zhou2024gate} assembles 5K human‐annotated instances across 56 real‐world tasks (e.g. travel guides, design ideation) with IntJudge to test open-ended multimodal generation methods on arbitrary instruction‐driven interleaved generation. In contrast, OpenLEAF~\cite{an2023openleaf} gathers 30 open‐domain queries with each written and reviewed by annotators to probe foundational interleaved text–image generation, measuring entity and style consistency via LMM evaluators plus human validation. Finally, MMIE~\cite{xia2024mmie} proposes a unified interleaved suite by sampling from 12 fields and 102 subfields, offering a mix of multiple-choice and open-ended question formats to evaluate models in a diverse manner. In a more recent work, UniBench~\cite{unieval} was introduced as a comprehensive compositional benchmark for evaluating unified models, offering 81 fine-grained tags to ensure high diversity.

\subsection{Evaluation on Unification}

\textcolor{black}{
The above evaluation paradigms mainly assess understanding and generation capabilities in isolation, which is insufficient to determine whether unified models can achieve a complementary interaction between understanding and generation tasks. For example, unified models may leverage their understanding ability to enhance generation, or use generative simulation to support more complete understanding.
}

\textcolor{black}{
To fill this critical gap, RealUnify~\cite{shi2025realunify} builds the evaluation on unification around two core aspects: (1) Understanding-Enhanced Generation (UEG), which requires reasoning (e.g., commonsense, logic) to guide image generation; and (2) Generation-Enhanced Understanding (GEU), which requires mental simulation or reconstruction (e.g., of transformed or disordered visual inputs) to solve reasoning tasks. RealUnify contains 1,000 carefully human-annotated instances, covering 10 categories and 32 subtasks. This benchmark combines direct end-to-end evaluation with diagnostic step-by-step evaluation, decomposing tasks into separate understanding and generation stages, enabling us to precisely determine whether performance bottlenecks arise from deficiencies in core capabilities or from failures to effectively integrate these capabilities.
}

% \subsection{Evaluation on Other Types of Benchmarks}

% Beyond unified understanding, image generation, and interleaved tasks, additional benchmarks probe large-scale conditional and personalized synthesis. MultiGen-20M~\cite{qin2023unicontrol} comprises over 20 million image–prompt–condition triplets sourced from LAION-Aesthetics-V2~\cite{schuhmann2022laionAestheticsV2}, enabling comprehensive automated evaluation of alignment under varied visual conditions and the the evaluation set with 100-300 imagecondition-prompt triplets for each task.

% DreamBench~\cite{ruiz2023dreambooth} introduces a personalized generation test spanning 30 reference objects paired with curated prompts and human-annotated fidelity judgments. DreamBench++~\cite{peng2024dreambench++} extends this framework to 150 diverse reference images and 1,350 prompts, employing advanced multimodal language models for automated, human-aligned scoring across concept preservation, compositional fidelity, and stylistic consistency. Together, these datasets offer a coherent spectrum from massive automated benchmarks to focused, human-centric assessments of conditional and subject-driven image generation.

%% file: sec/6_challenges.tex
\section{Challenges and Opportunities on Unified Models}
Currently, at its rudimentary stage, unified multimodal models face several significant challenges that should be addressed to achieve robust and scalable understanding and generation capabilities. 
First, the high dimensionality of visual and textual data leads to extremely long token sequences. Efficient tokenization and compression strategies are essential to reduce memory and computation costs while preserving representational fidelity. 
Second, cross-modal attention becomes a performance bottleneck as image resolution and context length increase. Scalable alternatives such as sparse or hierarchical attention mechanisms  may potentially mitigate this issue. 
Third, pretraining datasets often include noisy or biased image–text pairs, particularly for complex image compositions and interleaved image-text data. Reliable data filtering, debiasing, and synthesizing are crucial to ensure fairness and robustness. 
Fourth, evaluation protocols are typically designed for single tasks in isolation. There is a growing need for comprehensive benchmarks that assess both understanding and generation in an integrated manner, especially for sophisticated tasks such as image editing and interleaved image-text generation. 
Apart from the issues and challenges on architecture, data, and evaluation, applying chain-of-thought (CoT) reasoning and reinforcement learning (RL) techniques into unified MLLM models to improve both interpretability and performance \cite{jiang2025t2i} is also worth exploring. CoT can guide the model to generate intermediate reasoning steps, which are particularly beneficial for complex visual-question answering or image-conditioned generation. Meanwhile, RL can be used to optimize long-horizon objectives such as factual consistency, user satisfaction, or task success rate beyond token-level likelihoods.
Moreover, exploring the demographic and social biases of existing unified MLLM models \cite{liu2025fairness} is an important topic to ensure responsible deployment. As these models become increasingly capable across diverse modalities and tasks, unintentional amplification of cultural stereotypes, gender bias, or geographic imbalances embedded in pretraining data may result in harmful outputs. Future work should investigate effective fairness-aware training pipelines.
Finally, enabling personalized knowledge-driven generation within unified MLLMs \cite{an2025unictokens} is an emerging and important direction. Personalized models aim to incorporate user-provided concepts—such as specific objects, characters, or styles—into the model’s understanding and generation capabilities. However, current approaches often treat understanding and generation separately, using distinct concept embeddings for each task. This separation limits the model’s ability to generalize to compositional prompts that require implicit knowledge, such as generating "$<$\textit{bo}$>$ wearing its hat" without explicitly describing the hat. Unifying personalized understanding and generation under a shared modeling framework would allow better semantic grounding and contextual generalization. 
%Recent efforts such as UniCTokens demonstrate the feasibility of this approach by introducing unified concept tokens and progressive training strategies that align both tasks. Future unified MLLMs should further explore how shared representations and training objectives can support scalable and efficient personalization, enabling models to learn user-specific concepts once and apply them broadly across diverse multimodal tasks.
% add thinkdiff

To the best of our knowledge, most of current unified multimodal models primarily emphasize image understanding and text-to-image generation, while capabilities such as image editing are only attained through post-finetuning. Moreover, advanced functionalities like spatially controlled image generation, subject(s)-driven image generation, and interleaved image-text generation remain largely unexplored in the unified framework. Consequently, we believe there are abundant opportunities to advance the field by addressing key areas such as architectural design, training efficiency, dataset curation, evaluation methodologies, fairness, and reasoning to achieve unified multimodal models.

%% file: sec/7_conclusion.tex
\section{Conclusion}

We have presented a comprehensive view on unified multimodal models that integrate vision–language understanding and image generation within a single framework. Initially, we provide a concise overview of the foundational knowledge and recent advancements in both multimodal understanding and text-to-image generation models. Subsequently, we systematically survey unified multimodal models by categorizing them into three main paradigms: diffusion-based, autoregressive-based, and hybrid-based approaches. For each category, we introduce related works and further subdivide them into distinct subcategories to help readers better grasp the landscape of this field. Additionally, we curate relevant datasets and benchmarks to facilitate practical implementation and evaluation. Finally, we discuss the key challenges and opportunities in this domain, emphasizing that the study of unified multimodal models is still in its infancy. We hope that our survey will serve as a valuable resource to advance research and innovation in the development of unified multimodal models.

% We have presented a unified view of large-scale multimodal models that combine vision–language understanding and image generation within a single framework.
% By surveying both diffusion and autoregressive approaches, we highlighted how different tokenization strategies and architectural designs enable models to process and generate across modalities. We also reviewed the major datasets and benchmarks that drive progress, from large-scale image–text corpora to specialized evaluation suites on core tasks the model requires. Moreover, we analyzed recent advances in hybrid architectures that fuse autoregressive generation with diffusion processes, demonstrating how these combined methods balance symbolic control and visual fidelity. Our taxonomy clarifies the landscape of unified models, categorizing them by encoding strategies—pixel-based, semantic, query-based, and hybrid—and by backbone paradigms, thus providing a structured roadmap for future research.

% Looking ahead, the primary challenges include developing efficient encodings for high-resolution imagery, achieving end-to-end alignment between semantic and pixel representations, and mitigating biases inherent in large-scale data. Extending unified architectures to support additional modalities such as audio, video, and 3D data promises to bring us closer to truly general generative intelligence.